\documentclass[letterpaper]{article} 
\usepackage[submission]{aaai24_}  
\usepackage{times}  
\usepackage{helvet}  
\usepackage{courier}  
\usepackage[hyphens]{url}  
\usepackage{graphicx} 
\usepackage{natbib}  
\usepackage{amsmath}
\usepackage{amssymb}
\usepackage{caption} 

\usepackage{xcolor}

\usepackage{algorithm}
\usepackage{algorithmic}
\usepackage{booktabs}
\usepackage{multicol}
\usepackage{multirow}
\usepackage{comment}
\usepackage{amsmath}

\usepackage[switch, displaymath, mathlines]{lineno}

\newcommand\XP[1]{\textcolor{black}{#1}}

\newcommand\ourmodel{Gramformer}

\usepackage{newfloat}
\usepackage{listings}
\DeclareCaptionStyle{ruled}{labelfont=normalfont,labelsep=colon,strut=off} 
\floatstyle{ruled}
\newfloat{listing}{tb}{lst}{}
\floatname{listing}{Listing}
\gdef\showauthors@on{T}
\def\theauthors{\if T\showauthors@on\@author\else Anonymous submission\fi}

\title{Gramformer: Learning Crowd Counting via Graph-Modulated Transformer\footnote{This is the accepted version of \cite{lin2024gramformer}, the Proceedings of the 38th AAAI Conference on Artificial Intelligence (AAAI24), February 20-27, 2023, Vancouver, CA. \\ Please cite the final published version.}}

\author{
    Hui LIN$^1$,
Zhiheng MA$^2$,
Xiaopeng HONG$^{3,4}$\thanks{Xiaopeng HONG is the corresponding author.}, Qinnan SHANGGUAN$^3$,
Deyu MENG$^1$,\\ 
    \textsuperscript{\rm 1}School of Mathematics and Statistics, Xi'an Jiaotong University\\
    \textsuperscript{\rm 2}Shenzhen Institute of Advanced Technology, Chinese Academy of Science\\
    \textsuperscript{\rm 3}Harbin Institute of Technology\\
    \textsuperscript{\rm 4}Peng Cheng Laboratory\\
    {\tt\small linhuixjtu@gmail.com; zh.ma@siat.ac.cn;  hongxiaopeng@ieee.org;}\\ {\tt\small sg12qt@gmail.com; dymeng@mail.xjtu.edu.cn}}

\usepackage{bibentry}

\begin{document}

\maketitle

\begin{abstract}
\emph{Transformer} has been popular in recent crowd counting work since it breaks the limited receptive field of traditional CNNs. However, since crowd images always contain a large number of similar patches, the self-attention mechanism in Transformer tends to find a homogenized solution where the attention maps of almost all patches are identical. In this paper, we address this problem by proposing Gramformer: a graph-modulated transformer to enhance the network by adjusting the attention and input node features respectively on the basis of two different types of graphs. Firstly, an attention graph is proposed to diverse attention maps to attend to complementary information. The graph is building upon the dissimilarities between patches, modulating the attention in an anti-similarity fashion. Secondly, a feature-based centrality encoding is proposed to discover the centrality positions or importance of nodes. We encode them with a proposed centrality indices scheme to modulate the node features and similarity relationships. Extensive experiments on four challenging crowd counting datasets have validated the competitiveness of the proposed method. Code is available at {https://github.com/LoraLinH/Gramformer}.
\end{abstract}

\section{Introduction}

Crowd counting, which aims to estimate the number of people in crowded scenes, is a core and challenging task in computer vision. It has a wide range of real-world applications, including congestion estimation, traffic monitoring, and other crowd management scenarios, especially after the outbreak of COVID-19. The increasing demand drives the prosperous development of counting methods.


\renewcommand{\tabcolsep}{6 pt}{
\begin{figure}[t!]
	\begin{center}
		\begin{tabular}{ccc}
			
			\includegraphics[width=0.3\linewidth]{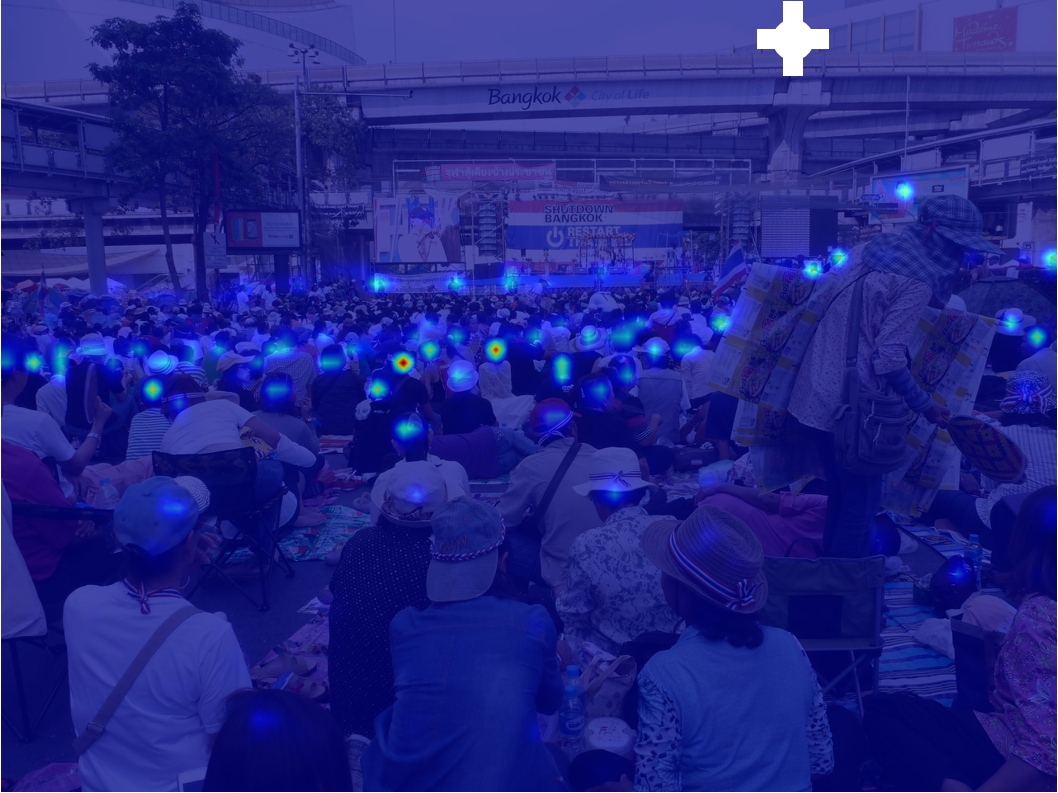}  &
			\includegraphics[width=0.3\linewidth]{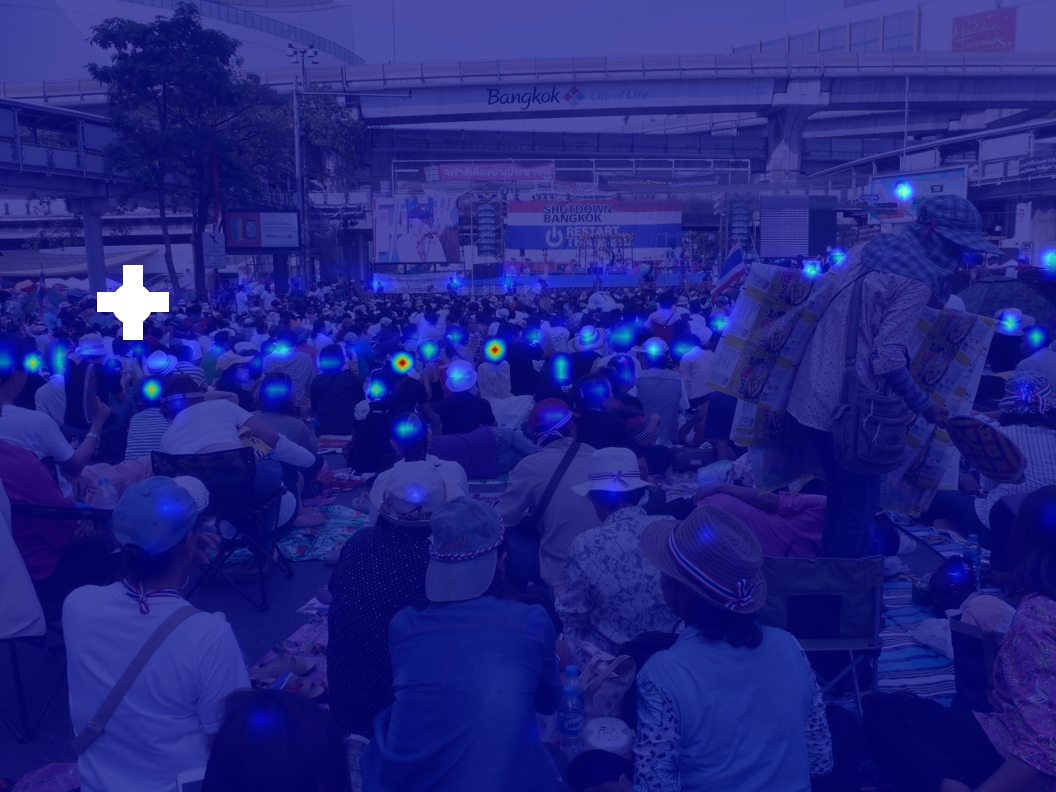} &
			\includegraphics[width=0.3\linewidth]{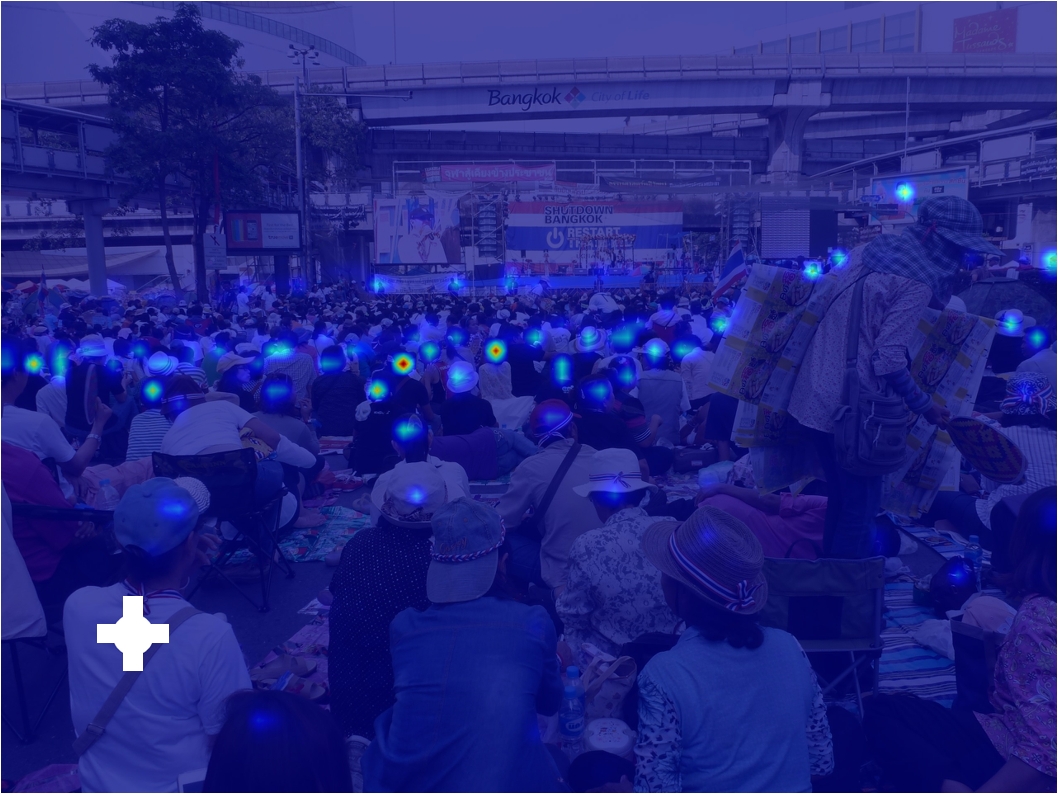} \\
			
			\multicolumn{3}{c}{\footnotesize{Vanilla Attention}}\\
			
			\includegraphics[width=0.3\linewidth]{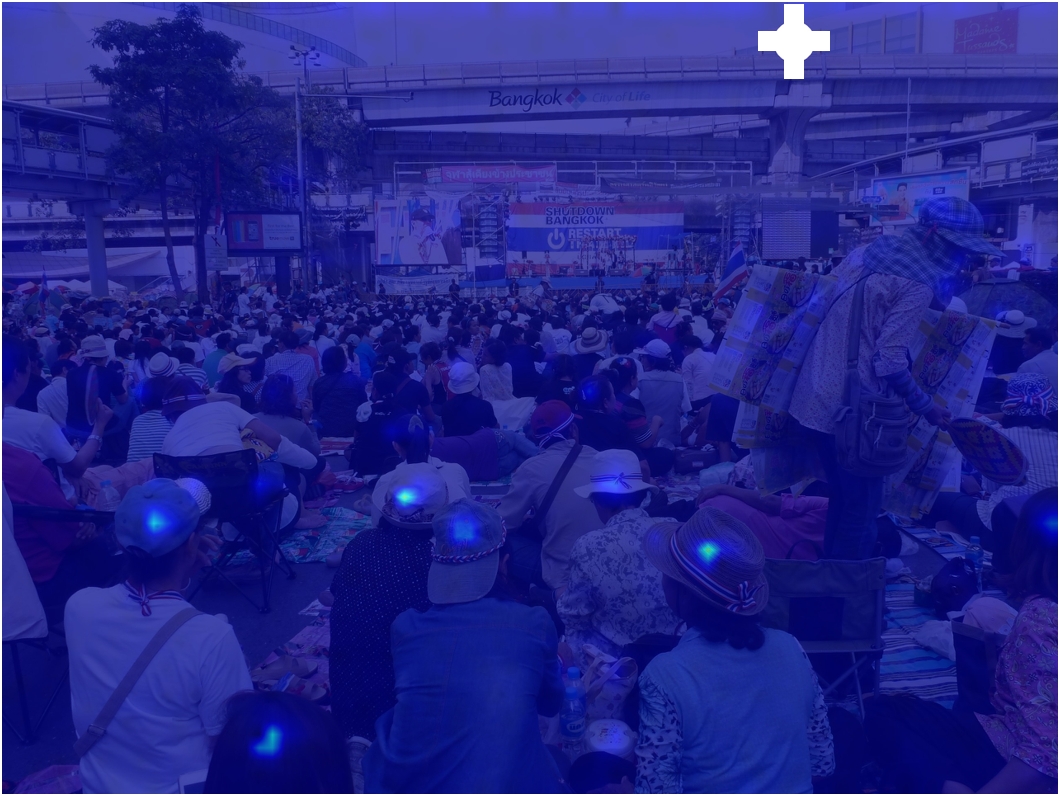}  &
			\includegraphics[width=0.3\linewidth]{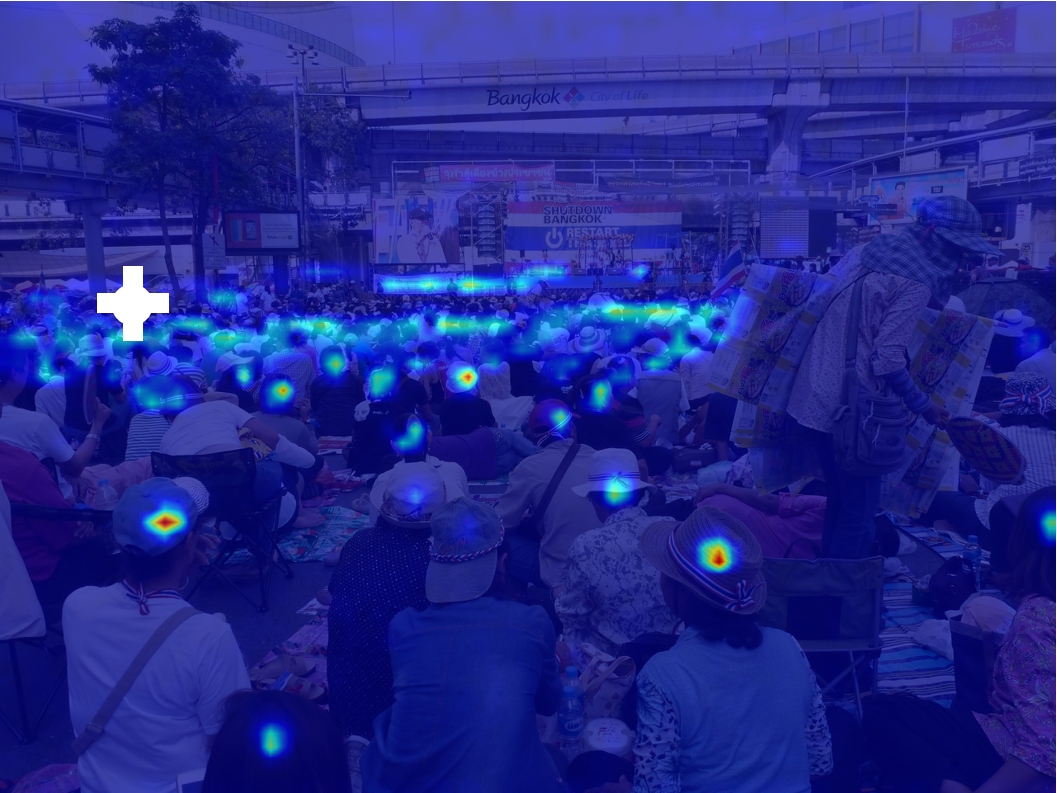} &
			\includegraphics[width=0.3\linewidth]{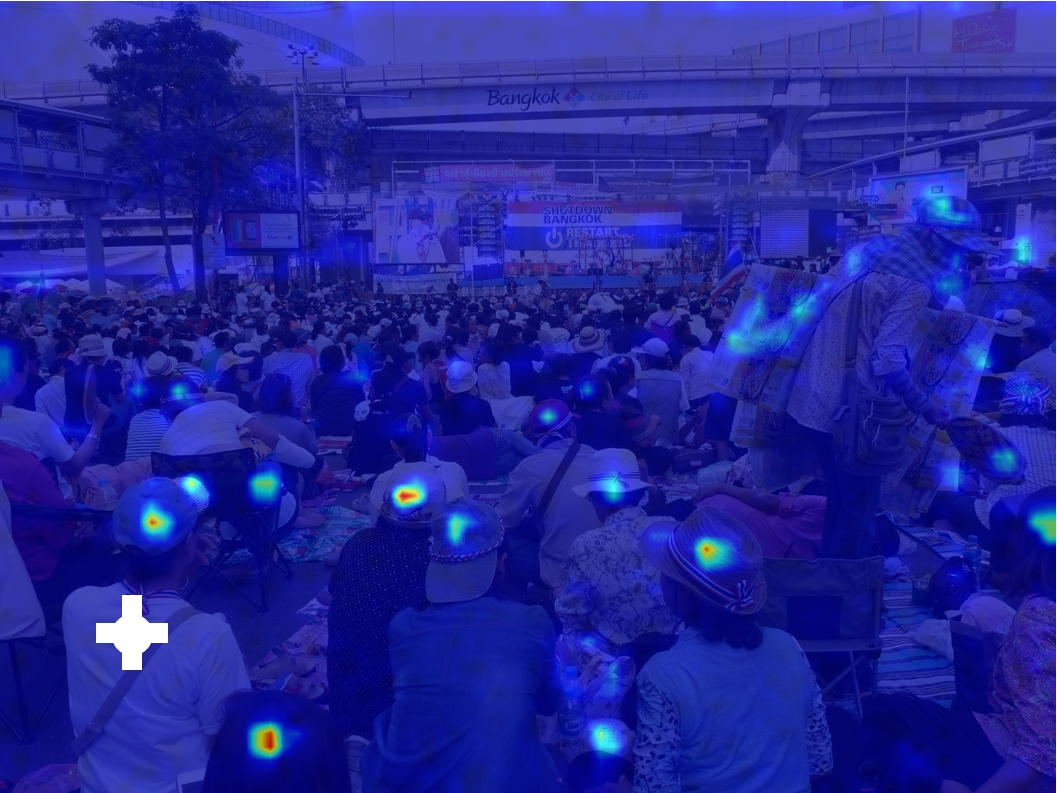} \\
			
			\multicolumn{3}{c}{\footnotesize{Graph-guided Attention}}\\

		\end{tabular}
		\vspace{-2mm}
		\caption{Visualization comparison of attention maps in different patches between vanilla attention and the proposed graph-guided attention. Each white cross mark indicates the location of a patch which the attention map is corresponding to. The vanilla attention finds a \XP{homogenized solution where the attention maps of most patches are similar (to the final density map)}, regardless of whether they are background (the first column) or foreground (the last two columns).}\label{fig:rational}
		
	\end{center}
\vspace{-6mm}
\end{figure}}

Over the last few decades, Convolution Neural Network (CNN) has been utilized as the mainstream structure to regress the density map and predict the total crowd count~\cite{zhang2016single, li2018csrnet, ma2019bayesian}. However, the accuracy of these methods will sometimes be constrained by the fixed-size convolution kernel, where each patch only receives information from its spatial neighbours. This makes traditional CNNs to be difficult to deal with different variations in crowd images. Recently, with the blossom of transformers, some approaches~\cite{liang2022transcrowd, lin2022boosting, lin2022semi} have leveraged on the self-attention mechanisms to further boost the accuracy. The transformer treats the image as a sequence of patches and each patch receives information from all other ones. With a dot-product operation, the intensity of information received is based on the feature similarity of different nodes. 

{While the transformer architecture is effective at computing crowd density in an image, it may produce homogenized \XP{attention maps} as there are numerous regions that appear similar in a crowdy image. In our experiments, we are surprised to find that the attention maps of most patches produced by the vanilla transformer are almost identical\footnote{The recent study~\cite{wang2023sclip} also reveals that the transformer in CLIP~\cite{radford2021learning} tends to produce highly similar attention maps across various source points in natural images.}, irrespective of whether they are foreground or background. As depicted in Figure~\ref{fig:rational}, these attention maps are predominantly focused on head positions and \emph{similar} to the crowd density map (\emph{homogenized attention}). It can be also seen that in Table~\ref{tab:identity}, the variance 
of the attention maps learned by the vanilla transformer on crowd images (QNRF) is \XP{below one-tenth of} those on natural images (ImageNet). The congruence observed between the homogenized attention maps and the density maps accelerates the production of the target density map by the counting model. However, it comes at the cost of overlooking other details that may be relevant. As a result, this scheme only focuses on the current attention region, ignoring its relationships with other patches, potentially leading to performance loss.

\renewcommand{\tabcolsep}{14 pt}{
\begin{table}[t]
\small
\vspace{-2mm}
\begin{center}
\begin{tabular}{ccc}
\toprule
Model & Dataset & \XP{ANVar}\\
\midrule
Vanilla Transformer & ImageNet val & 5028.2\\
Vanilla Transformer & QNRF test & 368.6\\
Our model & QNRF test & 4777.6\\
\toprule
\end{tabular}
\end{center}
\vspace{-2mm}
\caption{The comparison of the average normalized variance (ANVar) of attention maps generated by an Imagenet pre-trained transformer on natural images and crowded images (resized to a uniform size of 384$\times$384). The ANVar of the attention maps learned by the transformer on crowd images (QNRF) is significantly lower (\textless 10\%) than those on natural images (ImageNet). Our \ourmodel{} reduces homogenization and diversifies the attention maps.}\label{tab:identity}
\vspace{-4mm}
\end{table}}

\XP{In this paper, we address this problem by proposing a novel solution called \emph{graph-modulated transformer} (\emph{Gramformer} for crowd counting). We introduce two types of graphs, the attention graph and the feature-based centrality encoding graph, to modulate the attention and input node features of the transformer blocks, respectively. Building upon these two representations, we enhance the transformer network for crowd counting from two primary perspectives.}

\XP{First, we propose a graph-guided attention modulation method to diversify the generated attention maps. We suppose that the key to diverse attention maps lies in attending to complementary information. Based on this understanding, we design an Edge Weight Regression (EWR) network to encode the \emph{dissimilarity between patches} into edge weights. Considering the perspective geomery in crowd images, we further introduce an edge regularization term to limit} the dissimilarity encoded by EWR \XP{within the same horizontal line to be as little as possible. The graph edges form an attention mask, which is incorporated into the transformer block to modulate the attention maps in an anti-similarity fashion.}

\begin{figure*}[t]
\begin{center}
    \includegraphics[width=0.85\textwidth]{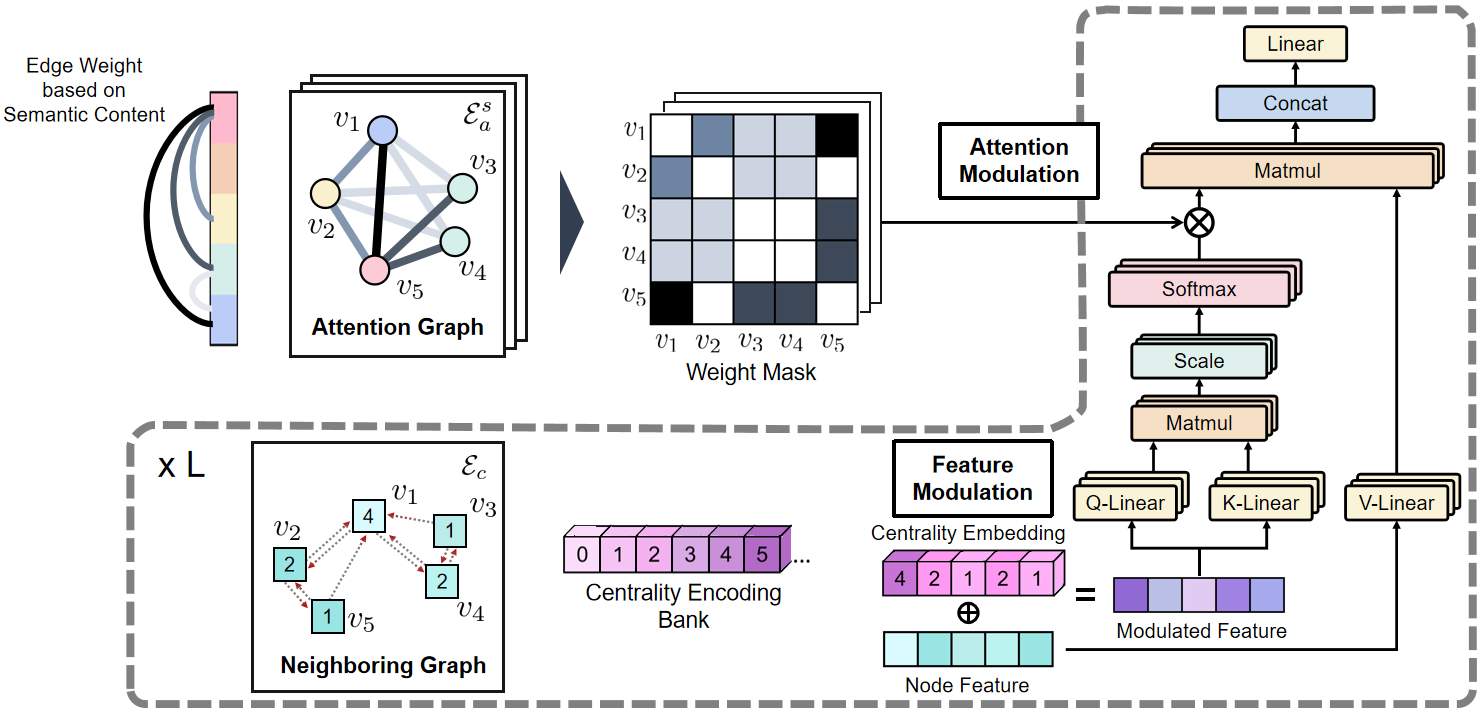}
\end{center}
\vspace{-2mm}
\caption{The framework of \ourmodel{}, which contains two main parts: an attention graph to modulate the attention mechanism by its edge weight, and a feature-based centrality encoding graph to encode the centrality or importance of each node. In the attention graph, different colors represent different semantic values predicted by EWR, and the color difference corresponds to the strength of connecting edges. In centrality encoding, \XP{each node is assigned a centrality index, which is linked to its in-degree. The centrality index is used to find the corresponding embedding from a learnable bank to modulate node features.}} 
\label{fig:structure}
\vspace{-4mm}
\end{figure*}

Secondly, we propose a graph-guided node feature modulation method that encodes the centrality positions or importance of nodes in the graph. To achieve this, we devise a \textit{Centrality Indices} scheme, which constructs a feature-based neighboring graph and explores the overall similarity of all nodes. Then we encode the graph structural information by the centrality embedding vectors to modulate input node features and thus further enhances the self-attention relations.




\XP{We propose a graph-modulated transformer architecture for crowd counting to modulate the attention and input node features of the transformer blocks, respectively. The contributions in detail can be summarized as follows:}

\begin{itemize}
    \item \XP{We propose a graph-guided attention modulation method to diversify the generated attention maps in an anti-similarity manner.}

    \item \XP{We propose an edge weight regression network with an edge regularization training term, to determine and regularize the edge weights for the \emph{attention graph}.}
    
    
    
    \item \XP{We propose a graph-guided node feature modulation method to capture centrality information and modulate the input feature node.}
    
	
    \item  Our approach consistently achieves promising counting performance on popular benchmarks.
	
\end{itemize}


\section{Related Works}

\subsection{Graph Transformer}
Recently, with the blossom of Transformer~\cite{vaswani2017attention}, researchers gradually focus on the generalization and combination between Transformers and GNNs~\cite{hanvision}. GraphTransformer~\cite{dwivedi2020generalization} introduces a generalization of transformer networks to homogeneous graphs. It incorporates the attention mechanism into graph structure and proposes to use Laplacian eigenvector as positional encoding. GraphTrans~\cite{wu2021representing} applies a permutation-invariant Transformer module after a standard GNN module. SAN~\cite{kreuzer2021rethinking} learns positional encodings by proposing the Laplace spectrum of a given graph. LSPE~\cite{dwivedigraph} presents a random-walk diffusion-based positional encoding scheme to initialize the positional representations of the nodes. It enables GNNs to learn both
structural and positional representations. Graphi~\cite{mialon2021graphit} encodes structural and positional information by positive definite kernels and local sub-structures such as paths of short length. GPS~\cite{rampavsek2022recipe} focuses and categorizes three main ingredients in unified transformer and graph neural networks: positional/structural encoding, local message-passing mechanism, and global attention mechanism. Graphormer~\cite{ying2021transformers} proposes several structural encodings to incorporate structural information reflected on nodes and the relation between node pairs. The proposed graph-guided transformer differs from existing approaches by modulating both the attention mechanism and node features respectively based on different types of constructed graph structures.

\subsection{Crowd Counting}
Early crowd counting methods adopt state-of-the-art detectors to perform the task, i.e., Faster R-CNN~\cite{ren2015faster}, YOLO~\cite{redmon2016you}, RetinaNet~\cite{lin2017focal}, etc. However, the detection-based methods~\cite{wu2005detection, idrees2013multi} appear to perform poorly in images with heavy occlusion. To improve this, density-based methods are introduced and gain promising counting accuracy. 

CNN is popularly used to generate predicted density maps since it possesses translation equivariance and is effective in extracting local details~\cite{liu2022convnet}, which has driven rapid developments in crowd countings~\cite{zhang2016single, li2018csrnet, ma2019bayesian}. Further improvements such as multi-scale mechanisms~\cite{zeng2017multi, ma2020learning, sindagi2017generating, cao2018scale, liu2019crowd}, perspective estimation~\cite{yang2020reverse, yan2019perspective, yang2020embedding}, auxiliary task~\cite{liu2020semi, zhao2019leveraging, yang2020weakly}, density refinement~\cite{liu2018crowd}, optimal transport~\cite{ma2021learning, lin2021direct, wang2020distribution, wan2021generalized} also show their effectiveness based on CNN.

The Vision Transformer (ViT)~\cite{dosovitskiy2020image} has demonstrated its outstanding performances in a variety of vision tasks~\cite{carion2020end, zheng2021rethinking, chen2021transformer}.
Lately, some works adopt transformer to boost the counting performance. TransCrowd~\cite{liang2022transcrowd} uses transformer to reformulate the weakly-supervised crowd counting problem from the perspective of sequence-to-count. MAN~\cite{lin2022boosting}incorporates global attention, learnable local attention, and instance attention into a counting model. CLTR~\cite{liang2022end} introduces an end-to-end transformer framework targeting on the crowd localization task. DACount~\cite{lin2022semi} proposes an agency-guided semi-supervised approach and uses a foreground transformer to refine crowd information.

Recently, HyGnn~\cite{luo2020hybrid} uses a Hybrid GNN to formulate information from different scales and domains. RRP~\cite{chen2020relevant} employs a vanilla GCN to propagate information between different regions. STGN~\cite{wu2022spatial} and CoCo-GCN~\cite{zhai2022co} target on video crowd counting and multi-view crowd counting respectively from graph perspective. Our method is distinct from them in two ways. First, we construct two different types of graphs based on the dissimilarity between each two patches and the overall similarity of all nodes. Second, these graphs will be leveraged to modulate the attention mechanisms with an anti-similarity fashion, and node features with a centrality indices scheme, respectively.



\section{The Proposed Method}


Given a crowd image $I$, we denote the feature map extracted by the backbone as $F \in \mathbb{R}^{C \times W \times H}$, where $C$, $W$, and $H$ are the channel, width, and height, respectively. These features can be represented as a set of nodes, denoted as $\mathcal{V} = \left\{v_1, v_2, ... , v_N \right\}$, where\footnote{Without being ambiguous, the symbol $v$ is also used to denote node features for simplicity.} $N=W \times H$.  We then build two graphs \XP{with two different ways of edge construction},
i.e., an attention graph with edges $\mathcal{E}_a$ and a feature-based centrality encoding graph with edges $\mathcal{E}_c$. Guided by these two graphs, we improve the transformer by modulating the attention mechanism using the edge weights produced by the attention graph and biasing the input node features based on the in-degree of the node in the neighboring graph, which is termed by the graph-modulated transformer (Gramformer).





\subsection{Structure of Graph-Modulated Transformer}



The structure of Gramformer is shown in Figure~\ref{fig:structure}. Gramformer's main features include an attention modulation scheme and a node feature modulation scheme. In details, given a single attention head  $s \in \{0,1,...,S\}$ and corresponding learnable matrices $W^s_Q, W^s_K \in \mathbf{R}^{C \times \frac{C}{S}}$, we define the graph-modulated multi-head attention as:
\begin{equation}
\label{eq:attn}
        \mathcal{R}^s(v^l_i, v^l_j) = E^s_{ij} \cdot \mathcal{S}(\frac{(\hat{v}^l_i W^s_Q) (\hat{v}^l_j W^s_K)^\mathsf{T}}{\sqrt{C}}).
\end{equation}
{$\mathcal{S}$ is the softmax function and $l$ is the transformer layer. $\hat{v}^l_i$ is the modulated node features, which is defined by}
\begin{equation}
\label{eq:posembedding}
        \hat{v}_i^l = v^l_i+\mathcal{P}(v^l_i).
\end{equation}

$E^s$ is a matrix of modulation factors, which is produced by the attention graph. $\mathcal{P}(v^l_i)$ is a biased vector, which is from a learnable embedding bank corresponding to the centrality index produced by the neighboring graph. Note that $E^s$ is set in the first layer of the transformer and kept constant for the remaining layers, while $\mathcal{P}(v^l_i)$ is re-computed in every layer of the transformer. In the following subsections, we will elaborate on how to calculate $ E^s$ and $\mathcal{P}(v^l_i)$ using two tailored graphs.

\subsection{Attention Graph}

The key to diverse attention maps lies in attending to complementary information. Based on this understanding, we propose an Edge Weight Regression (EWR) network to construct edges of the attention graph while encoding the dissimilarity between patches into edge weights.


Specifically, EWR estimates the semantic content value contained in each node and then determines the edges based on the degree of difference between the semantic values. For example, if the semantic content of nodes $v_A$ and $v_B$ have a large gap, a strong edge will be formed between $v_A$ and $v_B$ to encourage the model to explore the relationship between distinctive elements. On the other hand, if the node $v_C$ has a similar information content to the node $v_D$, the model will create a weaker or no connection between $v_C$ and $v_D$ to reduce excessive attention in the subsequent attention module. We also extend the EWR to multiple heads to construct heterogeneous graphs with different edge connections to encoder different kinds of semantics, which is consistent with multi-head attention. For head $s$, the weight and intensity of edge $E^s_{ij} \in \mathcal{E}_a$ between nodes $v_i$ and $v_j$ can be obtained by
\begin{equation}\label{eq:dip}
    E^s_{ij} = |\mathcal{F}^s_E(v^0_i) - \mathcal{F}^s_E(v^0_j)|.
\end{equation}
$\mathcal{F}^s_E$ denotes a head-specific EWR module, consisting of a feed-forward network (FFN) followed by an activating layer. We use a sigmoid function to normalize the output within a range of values from $0$ to $1$. In particular, $0$ prohibits an edge construction. We directly apply the edges to modulate the attention mechanism based on their weights and intensities, i.e., $E = \mathcal{E}_a$. EWR is in an anti-similarity fashion, diversifying the attention by appropriately reducing the intensity between weakly connected nodes. In this way, the attention is guided to be diversified, which can effectively discover the missing context and residual details. In our model, the attention graph will be generated by EWR after the backbone and remain constant throughout subsequent transformer layers.

We further introduce an Edge Regularization term to encourage the dissimilarity encoded by EWR within the same horizontal line to be as little as possible. The idea is motivated by 
the existence of the perspective geometry of pinhole cameras in crowd images, which usually generates perspective or scale variations only along the vertical axis~\cite{shi2019revisiting}. And within the same horizontal axis, these semantics remain consistent. Based on this observation, 
we propose a regularization to minimize the variance of the estimated semantic content value among nodes within the same horizontal axis. The regularization can be written as 
\begin{equation}
    \mathcal{Q} = \sum_h \ \sum_{\substack{i\\{\textit{s.t.}}\ y_i = h}} (\mathcal{F}_E(v^0_i) - \overline{\mathcal{F}}_h),
\end{equation}
where $y_i$ is the horizontal coordinate of each node and $\overline{\mathcal{F}}_h$ is the mean semantic value of the $h_{th}$ row.

\noindent \textbf{The Rationality of the Edge Regularization.}

Due to the imaging process of perspective geometry, crowd images often exhibit similar scales along the horizontal axis, resulting in similar density values in the predicted density map. As the vanilla transformer tends to produce attention maps highly similar to the predicted density map, as observed in the introduction, the attention map will also exhibit similar values along the horizontal axis, leading to excessive focus between different locations on the same horizontal axis. Overly homogeneous attention maps are detrimental to counting performance. Therefore, to diversify the attention maps generated, the proposed edge regularization penalty suppresses cases where similar attention values are produced along the same horizontal axis and breaks the downside of excessive attention.

\subsection{Feature-based Centrality Encoding}
\label{subsec:neighboring_graph}

We then go on to modulate the node features by looking at the overall structure and discovering the role or importance of each node. To achieve this, we devise a Centrality Indices scheme and a feature-based neighboring graph, which explores the overall similarity of all nodes and modulate the node features by encoding the centrality representation $\mathcal{P}$ into the node features.

For each node $v_i$, we select its similarity-based top $q$-NN, denoted as $\mathcal{N}(v_i)$ where $q$ is the percentage. We then construct directed graph edges $E^\prime_{ij} \in \mathcal{E}_c$ \XP{as follows:} 
\begin{equation}
    E^\prime_{ij} = \left\{
    \begin{aligned}
    &1, &\textup{if} \  v_j \in \mathcal{N}(v_i) \\
    &0, &\textup{otherwise}
    \end{aligned}
    \right..
\end{equation}
Here we adopt the L2 norm Euclidean distance to calculate the similarity. Then, we gather all the neighbour sets of the entire graph as $\mathcal{N} = \left\{\mathcal{N}(v_1), \mathcal{N}(v_2), ...,  \mathcal{N}(v_N) \right\}$. To identify the centrality, we count the number of occurrences of each node in $\mathcal{N}$. Nodes with high occurrences (in-degree in the graph) represent central positions in the feature space and we consider the occurrence as the centrality index. Suppose that $\left\{0, 1, ..., m \right\}$ are the set of centrality indices, where $m$ is the upper bound. All occurrences will be normalized to ensure that the maximum value does not exceed the bound.

We then introduce a learnable bank consisting of a series of centrality embedding vectors, denoted as $\left\{p_1,p_2,...,p_m \right\}$, $p \in \mathbb{R}^{C}$. Each node is encoded using the corresponding representation obtained through embedding matching. Specifically, we match the node $v^l_j$ with its centrality index $i$ from the bank, denoted as $\mathcal{P}(v^l_j) = p_i$, where $i$ corresponds to the occurrences of node $v^l_j$ within $\mathcal{N}$. In this way, we modulate features by encoding with distinctive vectors to incorporate the feature structural information into representations. 




\subsection{Further Processing}

Based on the attention map, the features will be further processed by:
\begin{equation}
    \Tilde{v}_i^l = LN(v^l_i + {\rm Concat}_s \left[\sum_{j=1}^N \mathcal{R}^s (\hat{v}^l_i, \hat{v}^l_j) \cdot (v^l_j W^s_v)\right]W_o),
\end{equation}
where $W^s_v \in \mathbf{R}^{C \times \frac{C}{S}}$ is the projection weight matrix and $\mathcal{R}^s$ is the graph-modulated attention map obtained by Equation~\ref{eq:attn}. The layer normalization (LN) function is used to balance the scales of the values. All information from $S$ graphs will be concatenated and finally integrated by the weight matrix $W^o \in \mathbf{R}^{C \times C}$ to update the node features. With an additional FFN (feedforward network) $\mathcal{K}$ to fuse the features, the updated nodes can be obtained by:
\begin{equation}
    \mathcal{V}^{l+1} = LN( \mathcal{\Tilde{V}}^{l} + \mathcal{K}(\mathcal{\Tilde{V}}^{l})).
\end{equation}
To be general, we denote the initial nodes extracted by the backbone as $\mathcal{V}^0$, and each transformer layer updates the node features to $\mathcal{V}^1, ..., \mathcal{V}^L$, where $L$ is the number of layers. 

It is important to note that while the attention graph $\mathcal{E}_a$ remains constant throughout the $L$ layers, the neighboring and similarity relationships will change in each layer after the update of features. This means that the feature-based neighboring graph is dynamic and the centrality indices in each layer will adjust to represent these changes. It allows the model more effectively and timely to capture the dynamic centrality feature positions of nodes.

Finally, we transmit graph nodes $\mathcal{V}^L$ into the density regression head to obtain the predicted density map. For counting supervision, we utilize the Instance Attention Loss (IAL) to mitigate the negative impact of annotation noise~\cite{lin2022boosting}. IAL dynamically focuses on the most important annotations and ignores supervisions with high uncertainty. Thus the final loss is expressed by
\begin{equation}
    \mathcal{L} = \mathcal{L}_{IA} + \lambda \mathcal{Q},
\end{equation}
where $\lambda$ is the weight of the regularization.


\section{Experimental Results}

\subsection{Datasets}

We evaluate our crowd counting method and compare it with other state-of-the-art methods on four largest crowd counting benchmarks. They are widely used in recent papers and are described as follows.  

\begin{itemize}
\item ShanghaiTech A~\cite{zhang2016single} includes 482 images with 244,167 annotated points. 300 images are divided for the training set and the remaining 182 images are for the test set. The numbers of people vary from 33 to 3,139.

\item UCF-QNRF~\cite{idrees2018composition} contains 1,535 crowded images with about 1.25 million annotated points. It has a wide range of people count and images with an average resolution of 2,013 $\times$ 2,902. The training and test sets include 1,201 and 334 images, respectively. 

\item NWPU-CROWD~\cite{wang2020nwpu} contains 5,109 images and 2.13 million annotated instances with point and box labels. 3,109 images are used in the training set; 500 images are in the validation set; and the remaining 1,500 images are in the test set. Images in NWPU-CROWD are in largely various density and illumination scenes. It is the largest dataset in which test labels are not released.

\item JHU-CROWD++~\cite{sindagi2020jhu} has a more complex context with 4,372 images and 1.51 million annotated points. 2,272 images are chosen for the training set; 500 images are for the validation set; and the rest 1,600 images for the test set. The dataset contains diverse scenarios and is collected under different environmental conditions of weather and illumination.
\end{itemize}


\def\arraystretch{1.1}
\renewcommand{\tabcolsep}{12 pt}{
\begin{table*}[t!]
\small
	\begin{center}
		\begin{tabular}{c|cc|cc|cc}
			\toprule[1.5pt]
			\multicolumn{1}{c}{Dataset} & \multicolumn{2}{c}{ShanghaiTech A} &  \multicolumn{2}{c}{UCF-QNRF} &  \multicolumn{2}{c}{JHU++} \\
			
			\multicolumn{1}{c}{Method} & \multicolumn{1}{c}{MAE} &
			\multicolumn{1}{c}{MSE} & \multicolumn{1}{c}{MAE} &
			\multicolumn{1}{c}{MSE} &  \multicolumn{1}{c}{MAE} &
			\multicolumn{1}{c}{MSE} \\
			\midrule
			MCNN~\cite{zhang2016single}\ (CVPR 16)  & 110.2 & 173.2 & 277 & 426 & 188.9 & 483.4\\
			CP-CNN~\cite{sindagi2017generating}\ (ICCV 17)  & 73.6 & 106.4 & - & - & - & - \\
			CSRNet~\cite{li2018csrnet}\ (CVPR 18)  & 68.2 & 115.0  &  - & - & 85.9 & 309.2 \\
			SANet~\cite{cao2018scale}\ (ECCV 18)  & 67.0  & 104.5 & - & - & 91.1 & 320.4\\
			CA-Net~\cite{liu2019context}\ (CVPR 19)  & 61.3 & 100.0 & 107.0  & 183.0 & 100.1 & 314.0 \\
		    CG-DRCN-CC~\cite{sindagi2020jhu}\ (PAMI 20)  & 60.2  & 94.0 & 95.5 & 164.3 & 71.0 & 278.6\\
		    DPN-IPSM~\cite{ma2020learning}\ (ACMMM 20)  & 58.1 & 91.7 & 84.7 & 147.2 & - & - \\
                BL~\cite{ma2019bayesian}\ (ICCV 19)  & 62.8 & 101.8 & 88.7 & 154.8 & 75.0 & 299.9 \\
		    DM-Count~\cite{wang2020distribution}\ (NIPS 20) & 59.7 & 95.7 & 85.6 & 148.3 & - & -  \\
		    UOT~\cite{ma2021learning}\ (AAAI 21)  & 58.1 & 95.9 & 83.3 & 142.3 & 60.5 & 252.7 \\		
			GL~\cite{wan2021generalized}\ (CVPR 21) & 61.3 & 95.4 & 84.3 & 147.5 & 59.9 & 259.5 \\	
                S3~\cite{lin2021direct}\ (IJCAI 21) & 57.0 & 96.0 & 80.6 & 139.8 & 59.4 & 244.0 \\
			P2PNet~\cite{song2021rethinking}\ (ICCV 21) & \textbf{52.7} & \textbf{85.1} & 85.3 & 154.5 & - & -  \\
             ChfL~\cite{shu2022crowd}\ (CVPR 22) & 57.5 & 94.3 & 80.3 & 137.6
            & 57.0 & 235.7 \\

            STEERER~\cite{han2023steerer} & 55.6 & 87.3 & \textbf{76.7} & 135.1  & 55.4 & 221.4 \\
			MAN~\cite{lin2021direct}\ (CVPR 22) & 56.8 & 90.3 & 77.3 & \underline{131.5} & \underline{53.4} & \textbf{209.9} \\
                CLTR~\cite{liang2022end}\ (ECCV 22) & 56.9 & 95.2 & 85.8 & 141.3 & 59.5 & 240.6 \\
                \ourmodel{} & \underline{54.7} & \underline{87.1} & \textbf{76.7} & \textbf{129.5} & \textbf{53.1} & \underline{228.1} \\
			\bottomrule[1.5pt]
		\end{tabular}
	\end{center}
 \vspace{-2mm}
\caption{Comparisons with the state of the arts on ShanghaiTech A, UCF-QNRF and JHU-Crowd++. The best performance is shown in \textbf{bold} and the second best is shown in \underline{underlined}. Note that the results of STEERER are based on the VGG19, which shares the same backbone as our method.}
\vspace{-2mm}
\label{tab:performance}
\end{table*}}

\def\arraystretch{1.1}
\renewcommand{\tabcolsep}{12 pt}{
\begin{table*}[t!]
\small
	\begin{center}
		\begin{tabular}{c|ccc|c|c}
			\toprule
			\multicolumn{1}{c}{Dataset}  & \multicolumn{3}{c|}{Overall} & \multicolumn{2}{c}{Scene Level (MAE)}\\
			 \multicolumn{1}{c}{Method} & MAE & MSE & NAE 
    &  Avg. & $S_0 \backsim S_4$ \\
			\midrule
		MCNN~\cite{zhang2016single} & 232.5 & 714.6 & 1.063 & 1171.9 & 356.0~/~72.1~/~103.5~/~509.5~/~4818.2 \\
		SANet~\cite{cao2018scale} & 190.6 & 491.4 & 0.991 & 716.3 & 432.0~/~65.0~/~104.2~/~385.1~/~2595.4 \\
		CSRNet~\cite{li2018csrnet} & 121.3 & 387.8 & 0.604 & 522.7 & 176.0~/~35.8~/~59.8~/~285.8~/2055.8 \\
		CAN~\cite{liu2019context} & 106.3 & 386.5 & 0.295 & 612.2 & 82.6~/~14.7~/~46.6~/~269.7~/~2647.0 \\
		BL~\cite{ma2019bayesian} & 105.4 & 454.2 & 0.203 & 750.5 & 66.5~/~8.7~/~41.2~/~249.9~/~3386.4 \\
		SFCN+~\cite{wang2020nwpu} & 105.7 & 424.1 & 0.254 & 712.7 & 54.2~/~14.8~/~44.4~/~249.6~/~3200.5 \\
	    UOT~\cite{ma2021learning} & 87.8 & 387.5 & 0.185 & 566.5 & 80.7~/~7.9~/~36.3~/~212.0~/~2495.4 \\
	    DMCount~\cite{wang2020distribution} & 88.4 & 388.6 & \underline{0.169} & 498.0 & 146.7~/~\underline{7.6}~/~\textbf{31.2}~/~228.7~/~2075.8 \\
	    
	    GL~\cite{wan2021generalized} & 79.3 & 346.1 & 0.180 & 508.5 & 92.4~/~8.2~/~35.4~/~\textbf{179.2}~/~2228.3 \\
     S3~\cite{lin2021direct} & 81.7 & 349.8 & 0.222 & 466.5 & 78.2~/~10.5~/~35.3~/~206.2~/~\underline{2002.4}\\
    ChfL~\cite{shu2022crowd} & 76.8 & 343.0 & 0.170 & 470.1  & 56.7~/~8.4~/~\underline{32.1}~/~195.1 ~/~2058.0 \\

    MAN~\cite{lin2022boosting} & \underline{76.5} & \underline{323.0} & 0.170 & \underline{464.6} & \underline{43.3}~/~8.5~/~35.3~/~190.9~/~2044.9 \\

	    \ourmodel{} & \textbf{72.5} & \textbf{316.4} & \textbf{0.160} & \textbf{441.9} & \textbf{26.9}~/~\textbf{7.4}~/~33.4~/~\underline{186.8}~/~\textbf{1955.2} \\
	    \toprule
		\end{tabular}
	\end{center}
 \vspace{-3mm}
\caption{Comparisons on NWPU-CROWD, which is further divided into \XP{five refined} subsets $S_0 \backsim S_4$ according to crowd size.} 
\vspace{-4mm}
\label{tab:performance2}
\end{table*}}

\subsection{Implement Details}

{\flushleft \textbf{Network Structure:}}
VGG-19 (Pretrained on ImageNet)~\cite{simonyan2014very} is adopted as the backbone network to extract features which will be formulated into graphs. The Edge Weight Regression (EWR) adopts two $3 \times 3$ convolution layers with a ReLU activation function in the middle and a sigmoid function at the end. The density regression head consists of an upsampling layer and three convolution layers with two ReLUs in middle. The kernel sizes of the first two layers are $3 \times 3$ and that of the last is $1 \times 1$.

{\flushleft \textbf{Training Details:}} 
We set the training batch size as $1$ and crop images with a size of $512 \times 512$. As some images in ShanghaiTech A contain smaller resolution, the crop size for this dataset changes to $256 \times 256$. Random scaling of $[0.75, 1.25]$ and horizontal flipping are also adopted to augment each training image. We use Adam algorithm~\cite{kingmaadam} with a learning rate $10^{-5}$ to optimize the parameters. We set the percentage of nearest neighbors $q$ as $30 \%$, and the maximum in-degree bound $m$ as $18$. The number of transformer layers $L$ is $2$ and the loss weight $\lambda$ is $0.1$.


\subsection{Comparison with state-of-the-art methods}

We evaluate our model on the above four datasets and list recent state-of-the-arts methods for comparison. The quantitative results, measured by MAE (Mean Absolute Error), MSE (Mean Square Error) and NAE (Normalized Absolute Error), are listed in  Tables~\ref{tab:performance} and~\ref{tab:performance2}.

Our method achieves great counting accuracy on all four benchmark datasets. Generally, by comparing \ourmodel{}{} against other state-of-the-art methods, \ourmodel{} outperforms or at least is on par with these competitors on all datasets. Specifically, on dense datasets NWPU and UCF-QNRF, our method achieves the best MAE and MSE. On JHU++, \ourmodel{} performs best on one of the evaluation metrics. And on the sparse dataset, ShanghaiTech A, \ourmodel{} ranks second and shows a competitive performance.

Furthermore, we observe that \ourmodel{} outperforms other approaches by large margins and is on par with the recent methods MAN and CLTR, which are built based on transformer network. We believe the characteristic of the attention module in the transformer works effectively on crowd counting task. And especially, it will help gain great accuracy improvements on dense datasets, i.e., UCF-QNRF, JHU++ and NWPU. Therefore, our graph-guided improvement in the application of attention to the counting task is an inspiring direction. We also present a visualization comparison in the appendix.

\subsection{Key Issues and Discussion}\label{sec:discussion}

\noindent \textbf{Ablation Studies}
We perform the ablation experiments on UCF-QNRF to study the effect of each proposed term in \ourmodel{} and provide quantitative results in Table~\ref{tab:contribution}.

\def\arraystretch{1.2}
\renewcommand{\tabcolsep}{8 pt}{
\begin{table}[t]
\small
\begin{center}
\begin{tabular}{lcc}
  \toprule[1pt]
  Components & MAE & MSE\\
  \midrule
  Baseline (VGG+IAL) & 84.7 & 150.9 \\
  B $+$ EWR & 79.3 & 135.4 \\
  B $+$ Centrality Indices & 79.8 & 138.1\\
  B $+$ EWR $+$ CI & 77.9 & 130.2\\
  B $+$ EWR $+$ Edge Regularization & 78.1 & 132.8 \\
  B $+$ EWR $+$ CI $+$ ER & 76.7 & 129.5 \\
  \toprule[1pt]
\end{tabular}
\vspace{-2mm}
\caption{Ablation study on UCF-QNRF. After combining all proposed components, the model achieves its best accuracy.}
\vspace{-3mm}
\label{tab:contribution}
\end{center}
\end{table}}

\def\arraystretch{1.1}
\renewcommand{\tabcolsep}{24 pt}{
\begin{table}[thbp!]
\small
\begin{center}
\begin{tabular}{c|cc}
\toprule
\multicolumn{1}{c}{$m$} & \multicolumn{1}{c}{MAE} & MSE\\
\midrule
 36 & 80.2 & 140.1\\
 24 & 78.4 & 136.9\\
 18 & \textbf{76.7} & \textbf{129.5}\\
 12 & 80.2 & 138.7\\
\toprule
\end{tabular}
\end{center}
\vspace{-4mm}
\caption{The influence of the number of centrality embedding vectors. Experiments are conducted on UCF-QNRF.}
\label{tab:encodings}
\vspace{-4mm}
\end{table}}

We start with the baseline of VGG+IAL, which details can be referred to MAN~\cite{lin2022boosting}. We first test the contribution of graph-modulated attention by EWR. The performance from baseline is improved by 5.4 and 15.5, for MAE and MSE, respectively. Then, we incorporate the central indices scheme into our model, the counting accuracy in terms of MAE is further improved by 3.9 without EWR and 1.4 with EWR. And finally, we adopt Edge Regularization to insert the prior knowledge of perspective geometry in crowd images. The model then achieves its best with $76.7$ and $129.5$ for MAE and MSE.

\noindent \textbf{Graph-guided Attention and Graph Transformer}
We \XP{study} the effectiveness of the proposed graph-guided attention. We choose Graphormer~\cite{ying2021transformers}, one of the most popular models among graph transformers, as our comparison model. We extract its attention module to replace our proposed graph-guided attention. Specifically, its edge connections are constructed according to the nearest neighbor similarities. The feature of each edge is obtained by concatenating the features of two endpoint nodes and smoothing with an MLP (Multilayer Perceptron). The feature of its edge is then encoded into a single value and added to the attention map, which is $R$ in Equation~\ref{eq:attn}. More detailed information about the model can be found in the appendix. The results of this baseline on UCF-QNRF is 80.9 for MAE and 142.5 for MSE, with a worse of 4.2 for MAE and 13.0 for MSE respectively. The comparison shows the efficacy of building edges by diversifying the attention maps \XP{for more abundant information}.

{\flushleft \textbf{The Influence of Centrality Indices:}}  We hold experiments on UCF-QNRF to study the influence of the number of centrality embedding vectors. $m$ is the upper bound of in-degree ranking values, determining the number of distinguished embedding vectors. The results are shown in Table~\ref{tab:encodings}. \XP{When $m=76.7$, our model performs best. When $m$ is reduced or elevated, the accuracy drops accordingly. We posit that an ill-suited value of $m$, resulting in either a limited number of encodings or an excessive set of embedding choices, has the potential to undermine the performance.}


{\flushleft \textbf{The Influence of The Percentage of Nearest Neighbors $q$:}} 
  $q$ determines how many nodes are selected as neighbors of a node when calculating its in-degree value.
A lower $q$ allows each node to select only the most similar ones, which reduces the number of nodes with high occurrences and corresponding centrality indices. And when $q$ grows larger, the neighbor selection process becomes more relaxed. At this point, nodes with high in-degree values will increase, influencing the learning ability of graph structure via embedding vectors. We hold experiments to study these influences. The results on UCF-QNRF are shown in Table~\ref{tab:neighbors}. \ourmodel{} achieves its best when $q=0.3$. And the accuracy drops when $q$ gradually increases or decreases. It justifies the importance of the nearest neighbor selection process in the updating of graph structure and subsequent count predictions. An appropriate number of selected neighbors will aid in improving the counting accuracy.

We also study the combined influence of $m$ and $q$. Nine models are trained with different settings: a cross-product of $m \in \{12, 18, 24\}$ and $q \in \{0.2, 0.3, 0.4\}$, to show a clear comparison. Table~\ref{tab:nnsize} reports MAE of counting results on UCF-QNRF. The counting accuracy of the model is affected by different settings. And \ourmodel{} performs best under the setting in the center with $m=18$ and $q=0.3$.

\def\arraystretch{1.1}
\renewcommand{\tabcolsep}{24 pt}{
\begin{table}[t]
\small
\begin{center}
\begin{tabular}{c|cc}
\toprule
\multicolumn{1}{c}{$q$} & \multicolumn{1}{c}{MAE} & MSE\\
\midrule
 0.06 & 81.6 & 140.1\\
 0.08 & 78.3 & 136.0\\
 0.1 & 78.8 & 135.4\\
 0.2 & 77.9 & 132.3\\
 0.3 & \textbf{76.7} & \textbf{129.5}\\
 0.4 & 78.2 & 133.1\\
\toprule
\end{tabular}
\end{center}
\vspace{-2mm}
\caption{The influence of the nearest neighbor selection.
When calculating the in-degree value, the decision of each node to connect to too many or too few ($q$) neighbors will reduce the counting accuracy.}

\label{tab:neighbors}
\vspace{-2mm}
\end{table}}

\def\arraystretch{1.2}
\renewcommand{\tabcolsep}{6 pt}{
\begin{table}[thbp!]
\small
\begin{center}
\begin{tabular}{r|ccc}
  \bottomrule
    Centrality & \multicolumn{3}{c}{Nearest Neighbors} \\
   Bank Size  & 0.2 & 0.3 & 0.4 \\
  \hline
   {12} & {\quad 78.8 \quad} & {\quad 80.2 \quad} & {\quad 81.5 \quad} \\
  {18} & {77.9} & {\textbf{76.7}} & {78.2} \\
  {24} & {81.4} & {78.4} & {80.6} \\
  \toprule
\end{tabular}
\end{center}
\vspace{-2mm}
\caption{Performances of \ourmodel{} under different settings of centrality embedding and nearest neighbors on UCF-QNRF (measured as MAE).}
\label{tab:nnsize}
\end{table}}

\renewcommand{\tabcolsep}{2 pt}{
\begin{table}[thbp!]
\small
\begin{center}
\begin{tabular}{l|ccc}
\toprule
\multicolumn{1}{c}{Model} & \multicolumn{1}{c}{Model Size (M)} & GFLOPs & Inference time \\
\midrule
ViT-B & 86.0 & 55.4 & 21.3 \\
Bayesian & 21.5 & 56.9 & 10.3\\
VGG19+Trans & 29.9 & 58.0 & 10.8 \\
MAN & 30.9 & 58.2 & 11.3\\
\textbf{\ourmodel{}}  & 29.0 & 60.9  & 12.6\\
    
\toprule
\end{tabular}
\end{center}
\vspace{-2mm}
\caption{\XP{Running Cost Evaluation.}}
\label{tab:parameter}
\vspace{-4mm}
\end{table}}

{\flushleft \textbf{Running Cost Evaluation:}} In Table~\ref{tab:parameter}, we compare the model size, the floating point operations (FLOPs) computed on one $384 \times 384$ input and the inference time for 100 images between \ourmodel{} and other models including of ViT-B~\cite{dosovitskiy2020image}, Bayesian~\cite{ma2019bayesian}, the combination of VGG19 and Transformer~\cite{vaswani2017attention} and  MAN~\cite{lin2022boosting}. All experiments are conducted with a single RTX 3080 GPU. \ourmodel{} has a slight increase in FLOPs and inference time,
\XP{for the computation of nearest neighbours in forward predictions}. However, our model is smaller than VGG19+Trans and MAN, and significantly smaller than ViT-B. This means that the proposed components are lightweight.

\section{Conclusion}
This paper aims to enhance the ability of transformers to modulate the attention mechanism and input node features respectively on the basis of two different types of graphs. We contribute to diversifying the attention map to attend to more complementary information by proposing a graph-guided attention modulation. We also encode the centrality or importance of nodes by designing a centrality indices scheme to adjust the input node features. The proposed \ourmodel{} achieves high counting accuracy on popular crowd counting datasets. Improving the transformer network by graph modulation is an inspiring direction, and we will apply it to a wider range of vision tasks.

\section{Acknowledgements}

This work is funded by the National Natural Science Foundation of China (62076195, 62206271, 62376070, 12226004 and 61721002), the Fundamental Research Funds for the Central Universities (AUGA5710011522), the Guangdong Basic and Applied Basic Research Foundation (2020B1515130004), and the Shenzhen Key Technical Projects under Grant (JSGG20220831105801004, CJGJZD2022051714160501, and JCYJ20220818101406014).

{\small
\bibliography{aaai24}
}

\clearpage

\section{Appendix}

\subsection{Detailed Structure of Graph Transformer Baseline}\label{sec:graphformer}

To demonstrate the importance of the proposed graph-modulated attention, we compare our Gramformer with the baseline structure constructed on the graph transformer (Graphormer~\cite{ying2021transformers}). Here we introduce the detailed structure of Graphormer for clearer analysis.  A depiction is shown in Figure~\ref{fig:graphormer}. Specifically, each node denotes a patch in the crowd image, and the edge connections are constructed according to the nearest neighbor similarities. For each node, it selects a certain number of nearest neighbors to connect a directed edge based on the feature similarities. Then the feature of each edge is obtained by concatenating the features of two endpoint nodes and smoothing with an MLP. In the graph transformer baseline, the edges will be encoded into weight values and added to the attention map. In particular, Given the top nearest neighbor set of each node as $\mathcal{N}(v_i)$, the weight of each edge can be expressed by:
\begin{equation}
    \Tilde{e}_{ij} = \left\{
    \begin{aligned}
    &\mathcal{Y}([v_i, v_j]), &\textup{if} \  v_j \in \mathcal{N}(v_i) \\
    &0, &\textup{otherwise}
    \end{aligned}
    \right.,
\end{equation}
where $\mathcal{Y}$ is an MLP module to predict the weight for each edge and $[\cdot,\cdot]$ denotes a concatenation of features.

Then the attention is obtained by adding the weight onto the dot-product matrix before the softmax operation. The attention of the graph transformer baseline is defined as
\begin{equation}
\label{eq:attn_final}
        \Tilde{\mathcal{R}}(v^l_i, v^l_j) = \mathcal{S}(\frac{(\hat{v}^l_i W_Q) (\hat{v}^l_j W_K)^\mathsf{T}}{\sqrt{C}} + \Tilde{e}_{ij}).
\end{equation}

For a fair comparison, we still use the centrality encoding to obtain the modulated input node feature $v^l$. Overall, our model has two significant differences from the baseline model. First, our edges are not constructed by relying on the nearest neighbour relationship, but are learned and generated by a regression network based on the perspective of diversifying attention among similar nodes. Second, our edge relations are applied in the attention as
\begin{equation}
\label{eq:attn2}
        \mathcal{R}(v^l_i, v^l_j) = e_{ij} \cdot \mathcal{S}(\frac{(\hat{v}^l_i W_Q) (\hat{v}^l_j W_K)^\mathsf{T}}{\sqrt{C}}.
\end{equation}
Compared to Equation~\ref{eq:attn_final}, the edge weights are directly applied after the softmax, and clearly reduce attention between nodes with similar characteristics, which have explicit implications.

\begin{figure}[t!]
\begin{center}
    \includegraphics[width=0.48\textwidth]{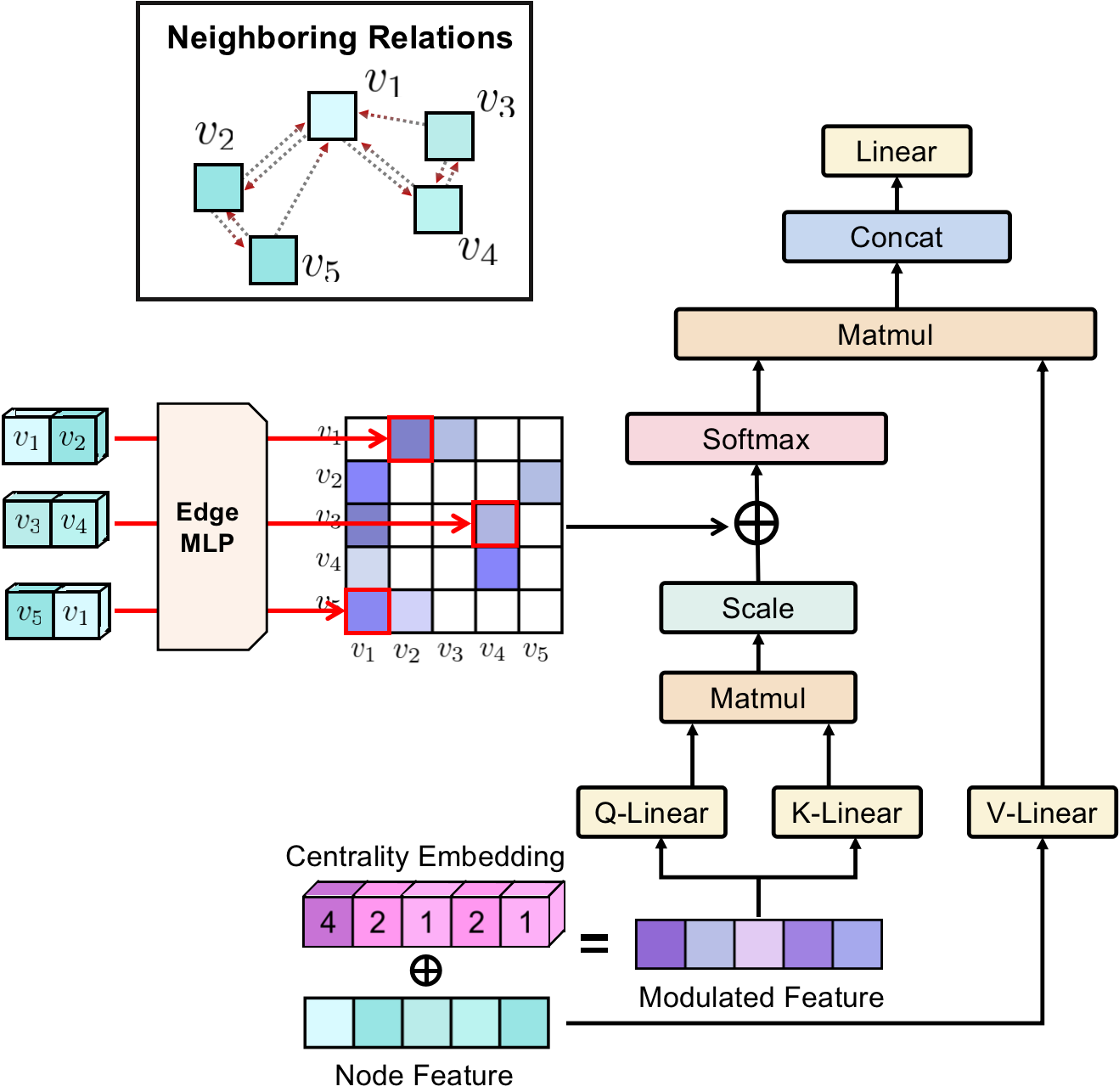}
\end{center}
\caption{Th overall structure of graph transformer baseline. The edge is constructed according to the nearest neighbor similarities and the weight, which is obtained by encoding the features of two endpoint nodes, will be added to the attention before softmax.}
\label{fig:graphormer}
\end{figure}

The results of this baseline on UCF-QNRF is 80.9 for MAE and 142.5 for MSE, with a worse of 4.2 for MAE and 13.0 for MSE respectively. The comparison result shows the effectiveness of building edges based on diversifying the attention maps to complementary information.

\subsection{The Visualizations of the Selected Nearest Neighbors in Neighboring Graph}\label{sec:visualizations}

In Figure~\ref{fig:neighbor}, we present a visualization of
the selected nearest neighbors for nodes in the neighboring graph. We sample five nearest neighbors for clarity. The red box represents the target node, while the other yellow boxes represent the nodes whose features are most similar to the target feature. 
The first row represents the neighboring relations of the features in the initial state before the transformer, while the second row represents the neighboring relations of the features after the transformer update.

\renewcommand{\tabcolsep}{12 pt}{
\begin{figure}[t!]
	\begin{center}
		\begin{tabular}{c}
			\includegraphics[height=0.55\linewidth]{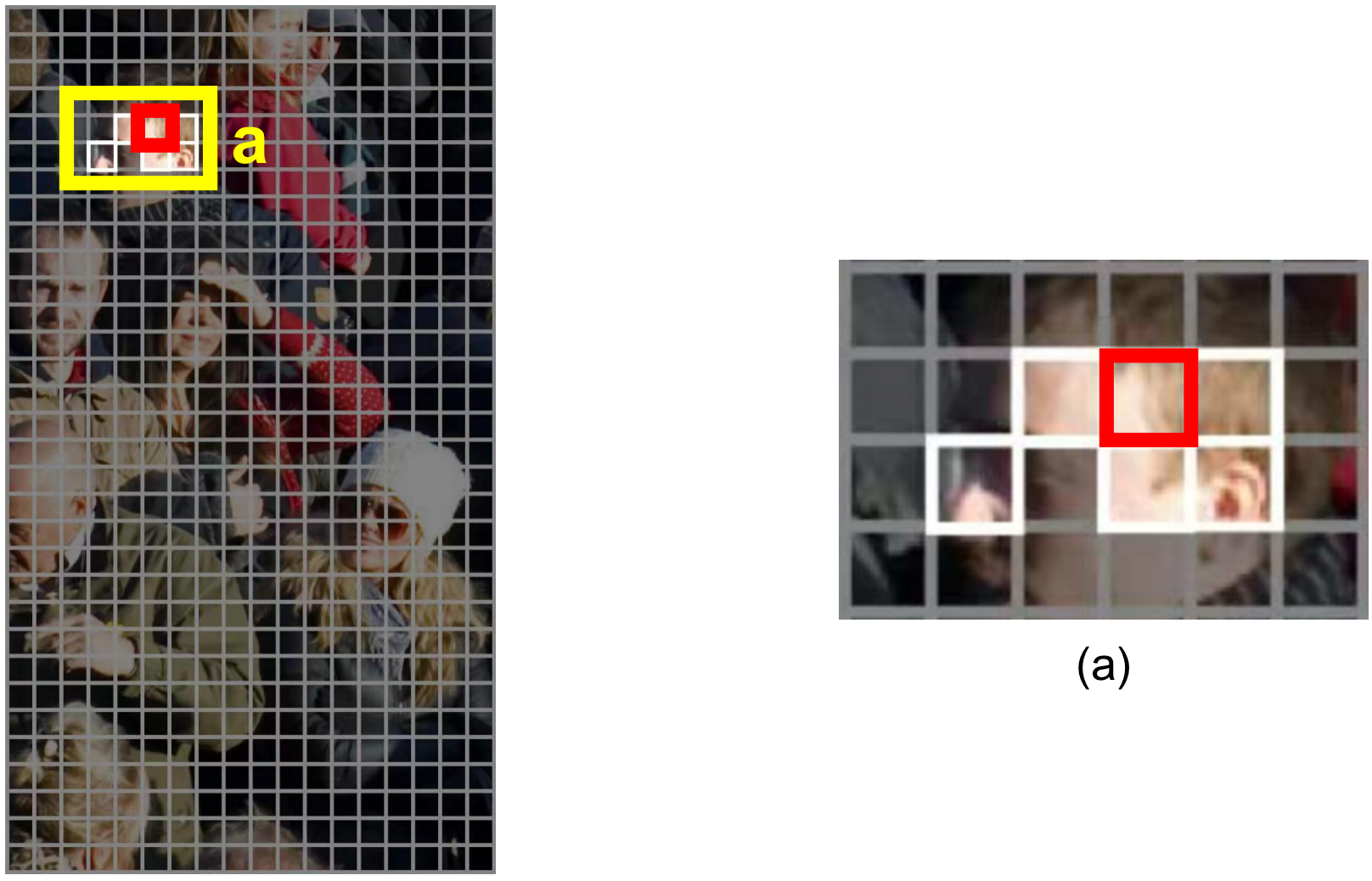}  \\
   
			\includegraphics[height=0.55\linewidth]{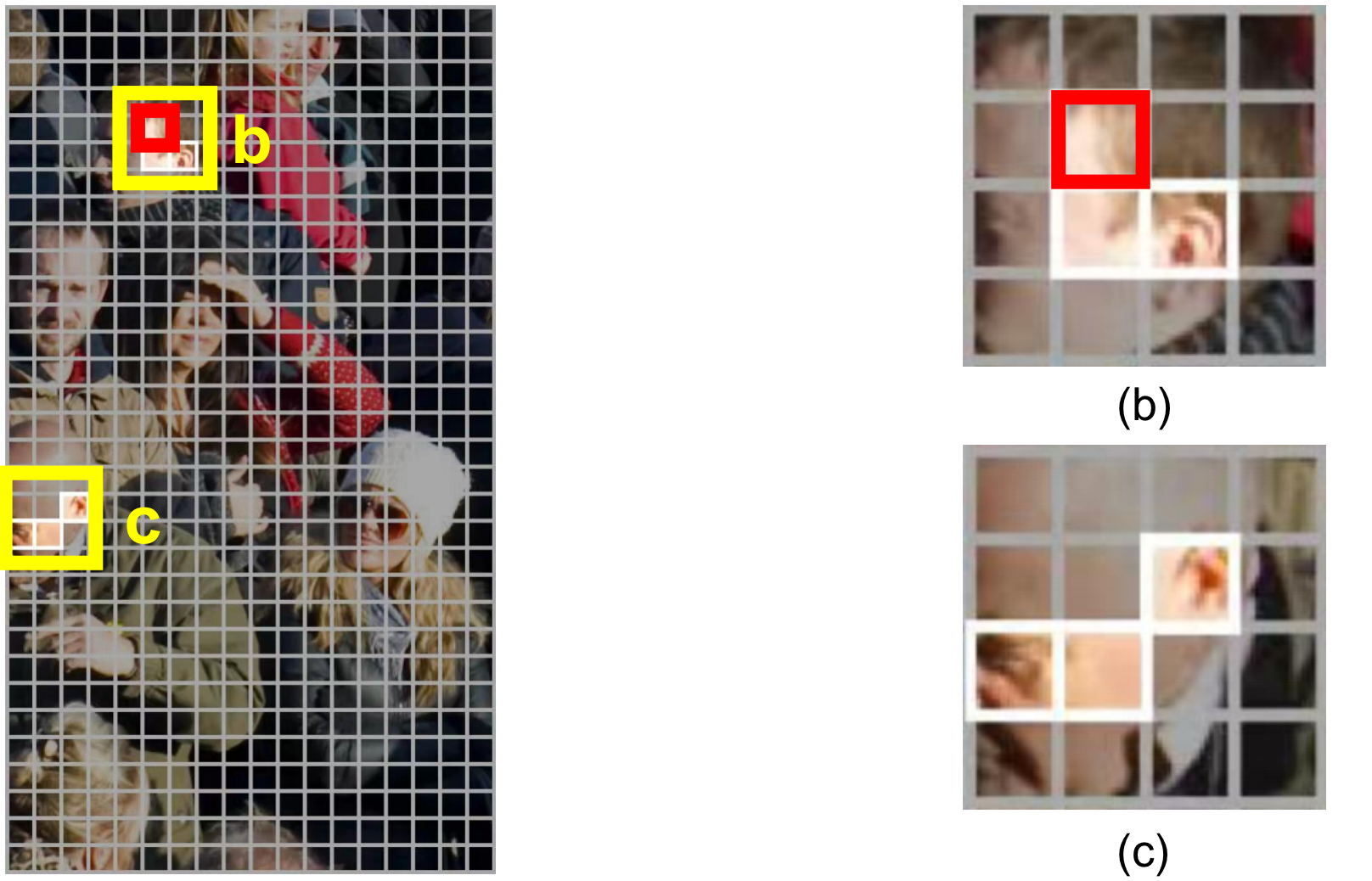}  \\
		\end{tabular}
		\caption{Visualizations of the selected nearest neighbors for nodes in the neighboring graph. The red box represents the target node, while the other yellow boxes represent the nodes whose features are most similar to the target feature. The first row represents the neighboring relations in the initial state, while the second row represents the neighboring relations after the transformer update.}
		\label{fig:neighbor}
	\end{center}
\end{figure}}

As can be seen, in the initial state, the neighboring nodes of the target are concentrated in its vicinity and belong to the same head. After the transformer update, the similarity of other heads with similar profiles is improved and the neighbors start to scatter to other nodes with similar profiles.
It proves that our model can effectively increase the feature similarity between nodes with similar profile and semantics.

\subsection{The influence of regularization weight $\lambda$}\label{sec:lambda}
We hold experiments to study the influence of the regularization weight $\lambda$, which controls the supervision intensity that encourages the dissimilarity encoded by EWR within the same horizontal line to be as little as possible. The results are shown in Table~\ref{tab:lambda}. All experiments are held on UCF-QNRF. It is noted that an inappropriate $\lambda$, i.e. 1, had a significant detrimental effect on the model's performance, even worse than not applying the regularization. 

\renewcommand{\tabcolsep}{12 pt}{
\begin{table}[h]
\small
\begin{center}
\begin{tabular}{|c|c|c|c|c|c|}
\hline
$\lambda$ & 0 & 0.001 & 0.01 & 0.1\\
\hline
MAE & 77.9 & 77.2 & 76.8 & 76.7\\
\hline
\end{tabular}
\end{center}
\caption{The influence of the regularization weight. An appropriate $\lambda$ during $[0.001, 0.1]$ leads to error reduction.}
\label{tab:lambda}
\end{table}}

\subsection{The influence of static attention graph and dynamic centrality encoding}\label{sec:encoding}
We hold experiments to study the influence of the static attention graph $E^s$ and dynamic centrality encoding $P(v_i^l)$. The setup is rooted in the module characteristics. We diversify the attention maps by encouraging them to monitor dissimilar features/regions for each transformer layer. A dynamic  attention graph can lead to extra differences in regions that were originally similar after updating, and their attention weights are untimely increased after regenerating the graph. This is a conflict with our intent and reduces the efficiency of diversity. On the other hand, the feature-based neighboring graph aims to encode the centrality information of each feature, which changes as the features are updated. Re-encoding the information at every layer enables the model to gain a deeper understanding of the similarity-based positional knowledge for these features. We hold the experiments and the results are shown in Table~\ref{tab:static}, where `S' and `D' denote static $E^s$ and dynamic $P(v^l_i)$ respectively.

\renewcommand{\tabcolsep}{10 pt}{
\begin{table}[thbp]
\small
\begin{center}
\begin{tabular}{|c|c|c|c|c|}
\hline
 & S+D (Ours) & S+S & D+D & D+S\\
\hline
MAE & 76.7 & 78.1 & 78.9 & 80.1\\
MSE & 129.5 & 131.7 & 136.2 & 140.5\\
\hline
\end{tabular}
\end{center}
\caption{The influence of static attention graph and dynamic centrality encoding. Unreasonable alternatives will result in a decrease in model's accuracy.}\label{tab:static}
\end{table}}

\subsection{The Visualizations of Predicted Crowd Density}\label{sec:crowd_visualizations}

We visualize the predicted density maps of proposed \ourmodel{} in Figure~\ref{fig:vis} and make comparisons with those predicted by the vanilla transformer. Upon observation, we notice distinct differences. In the first example (column), the vanilla transformer predicts blurred densities in crowd with smaller scales. In contrast, our model excels in predicting smaller crowd, resulting in remarkably clearer density location for each individual. In the subsequent two examples, our model also demonstrates its ability to accurately predict foreground densities, effectively avoiding false alarms that were evident in the background predictions generated by the vanilla transformer.

\renewcommand{\tabcolsep}{6 pt}{
\begin{figure*}[t!]
	\begin{center}
		\begin{tabular}{cccc} 
			\includegraphics[height=0.18\linewidth]{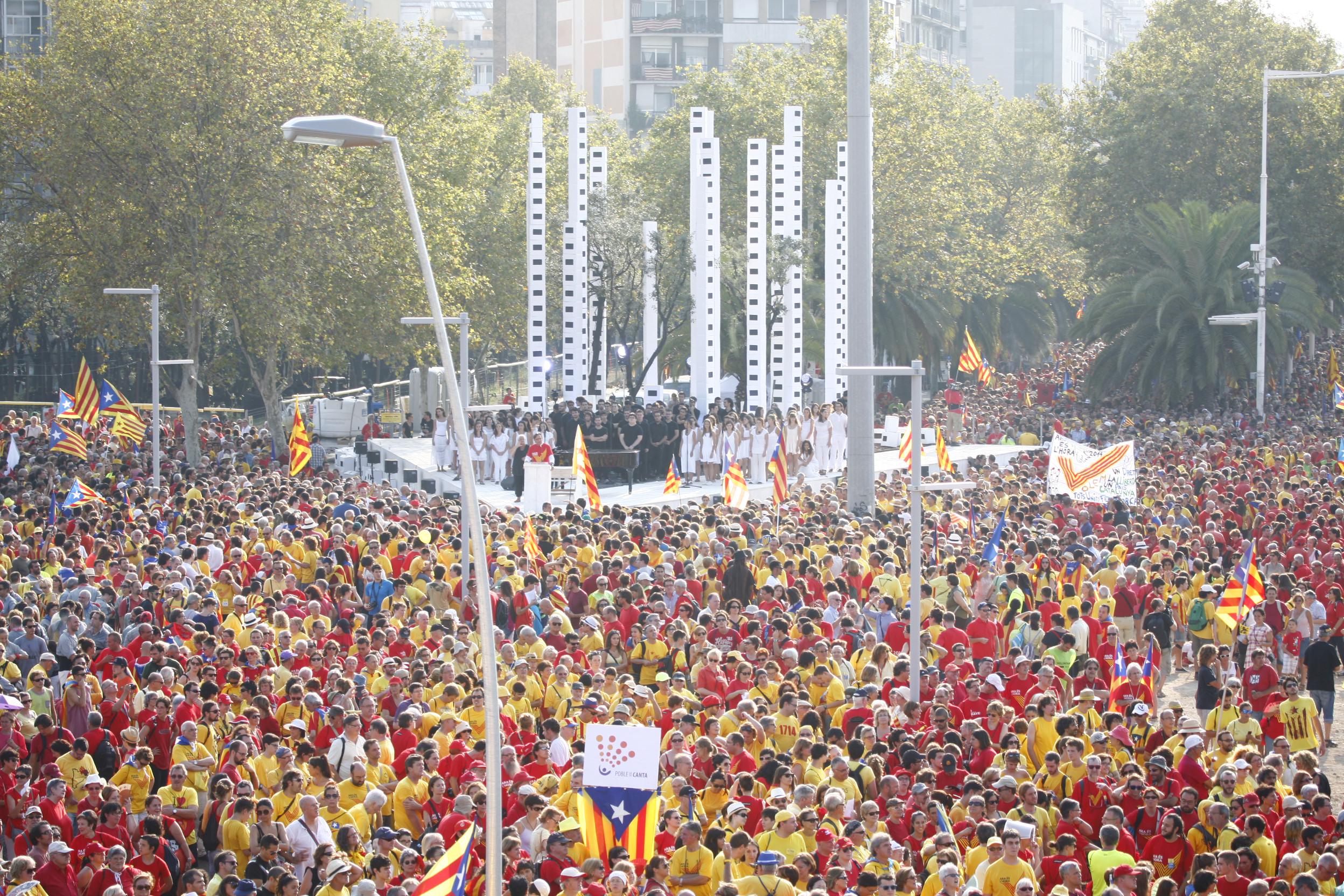}  &
			\includegraphics[height=0.18\linewidth]{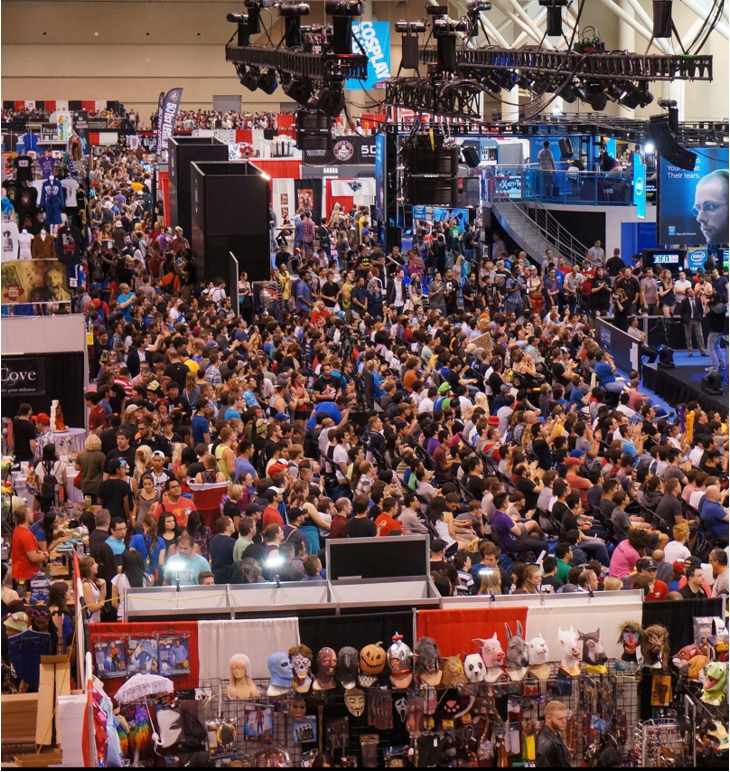} &
			\includegraphics[height=0.18\linewidth]{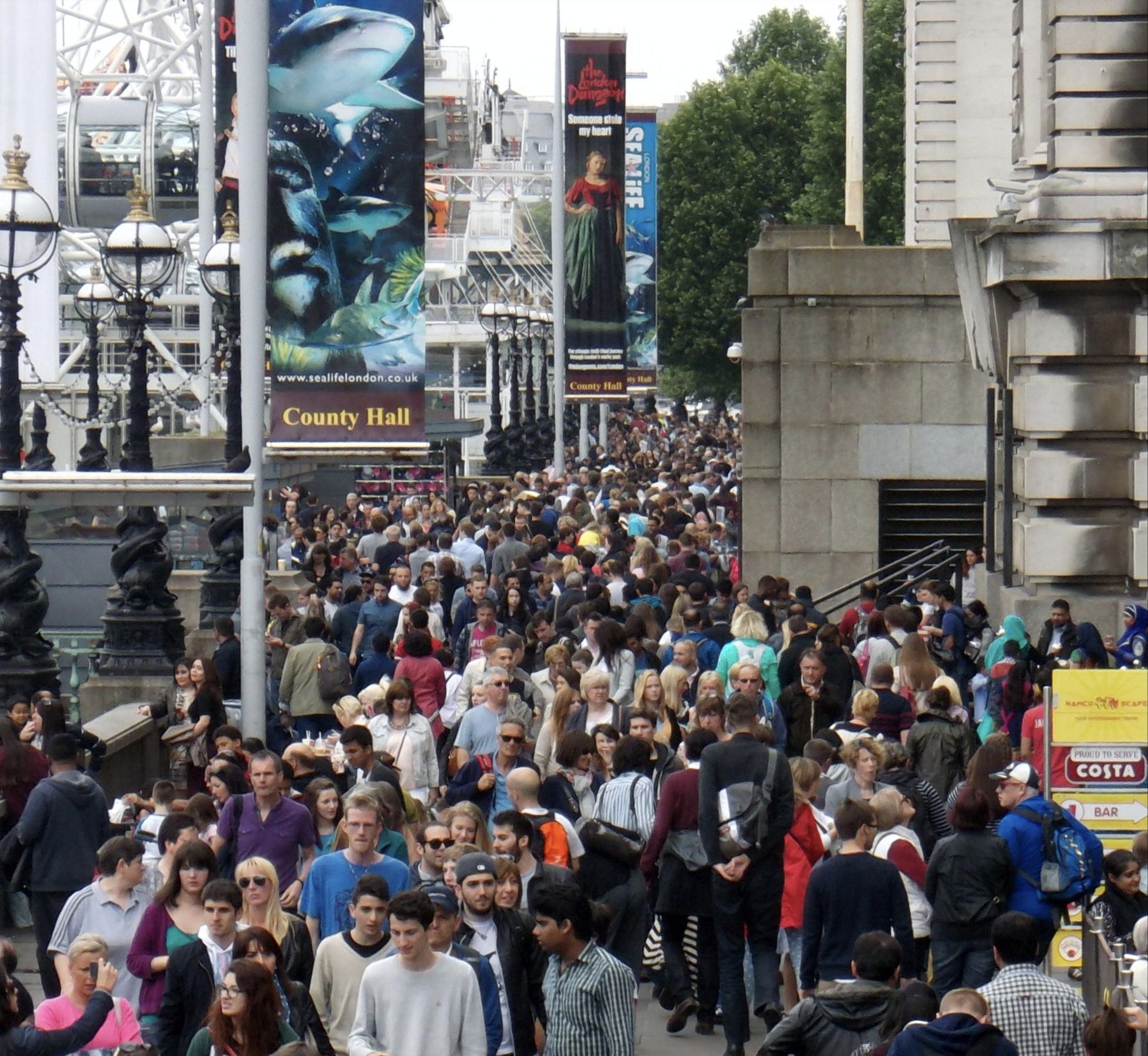} &
			\includegraphics[height=0.18\linewidth]{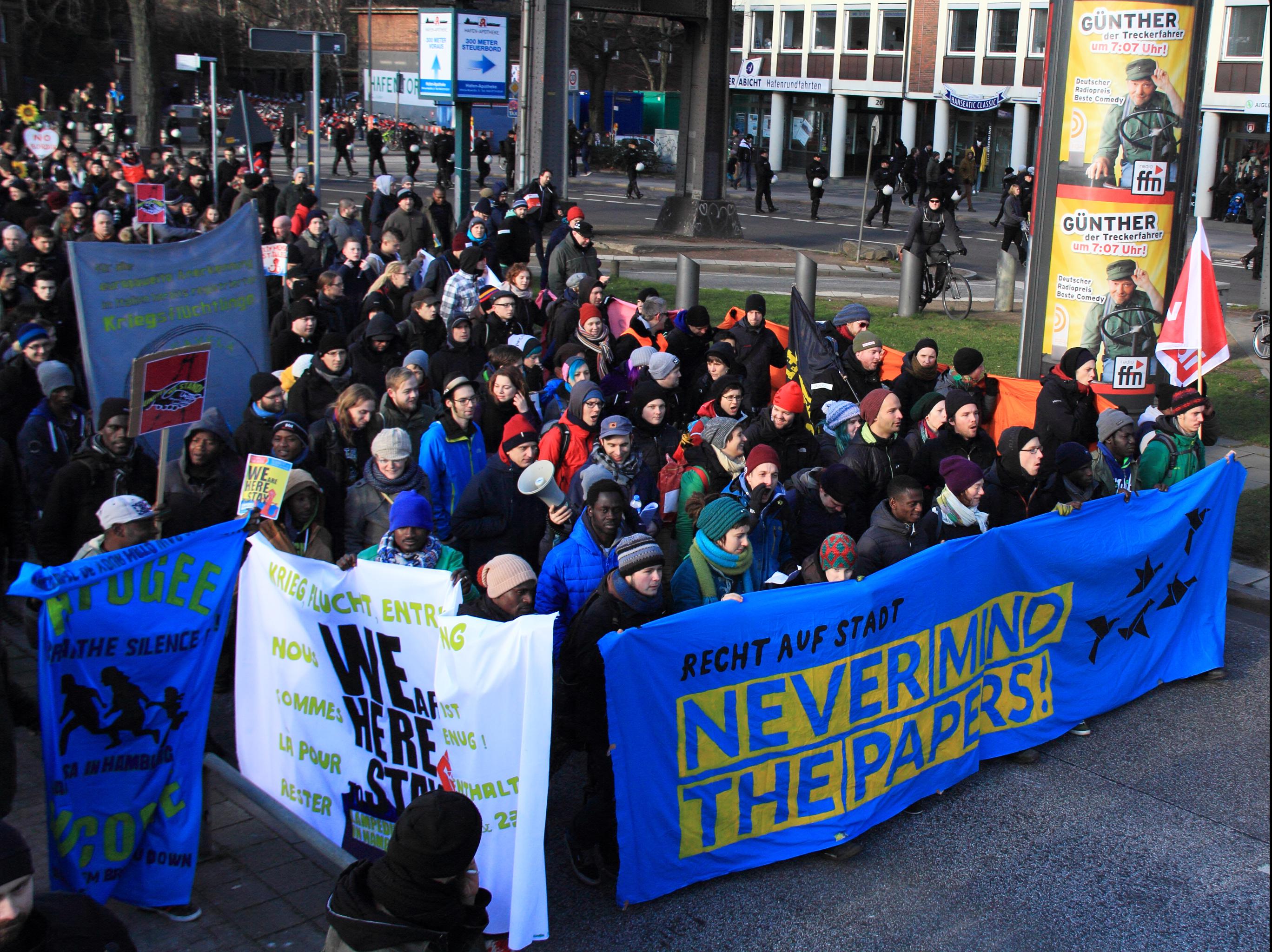} \\

			\includegraphics[height=0.18\linewidth]{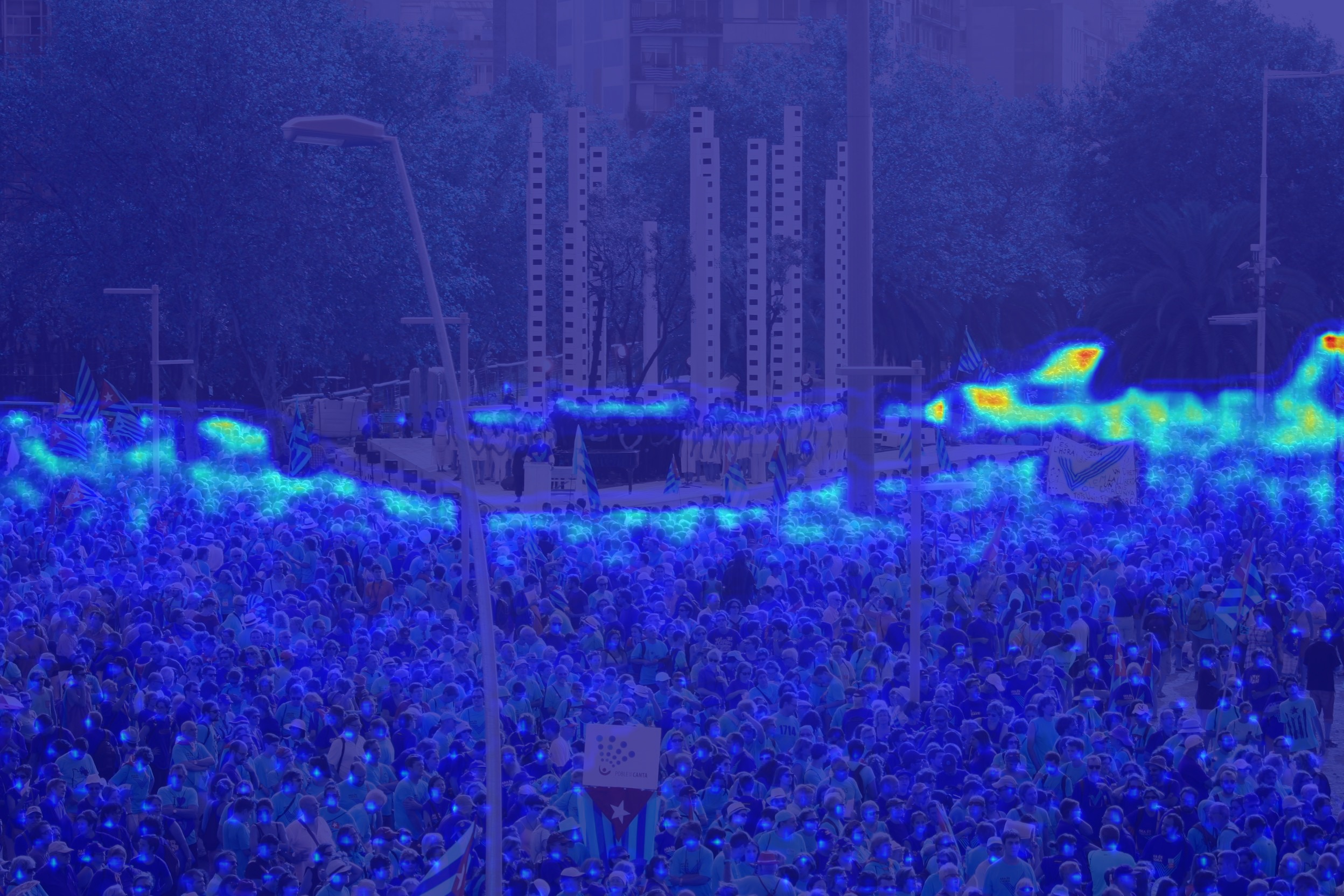}  &
			\includegraphics[height=0.18\linewidth]{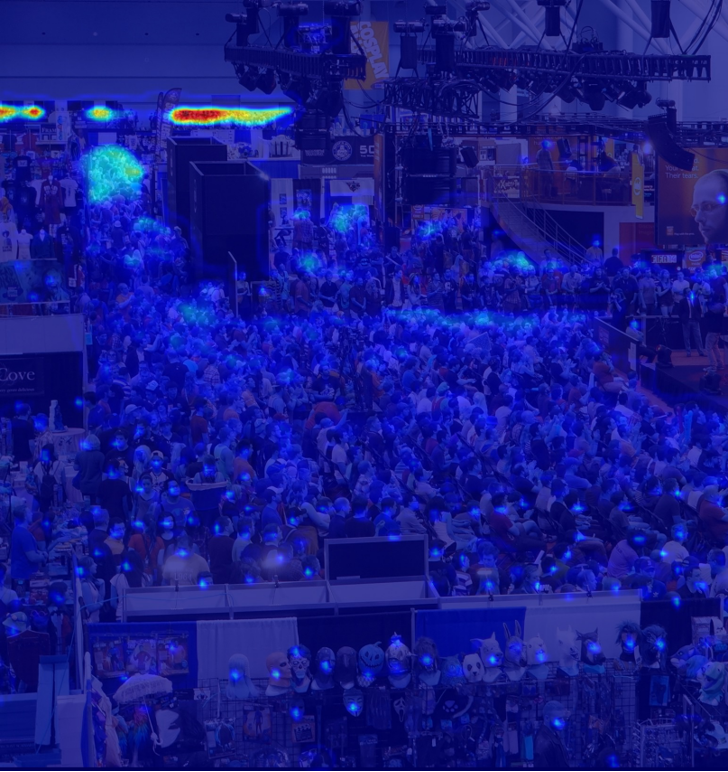} &
			\includegraphics[height=0.18\linewidth]{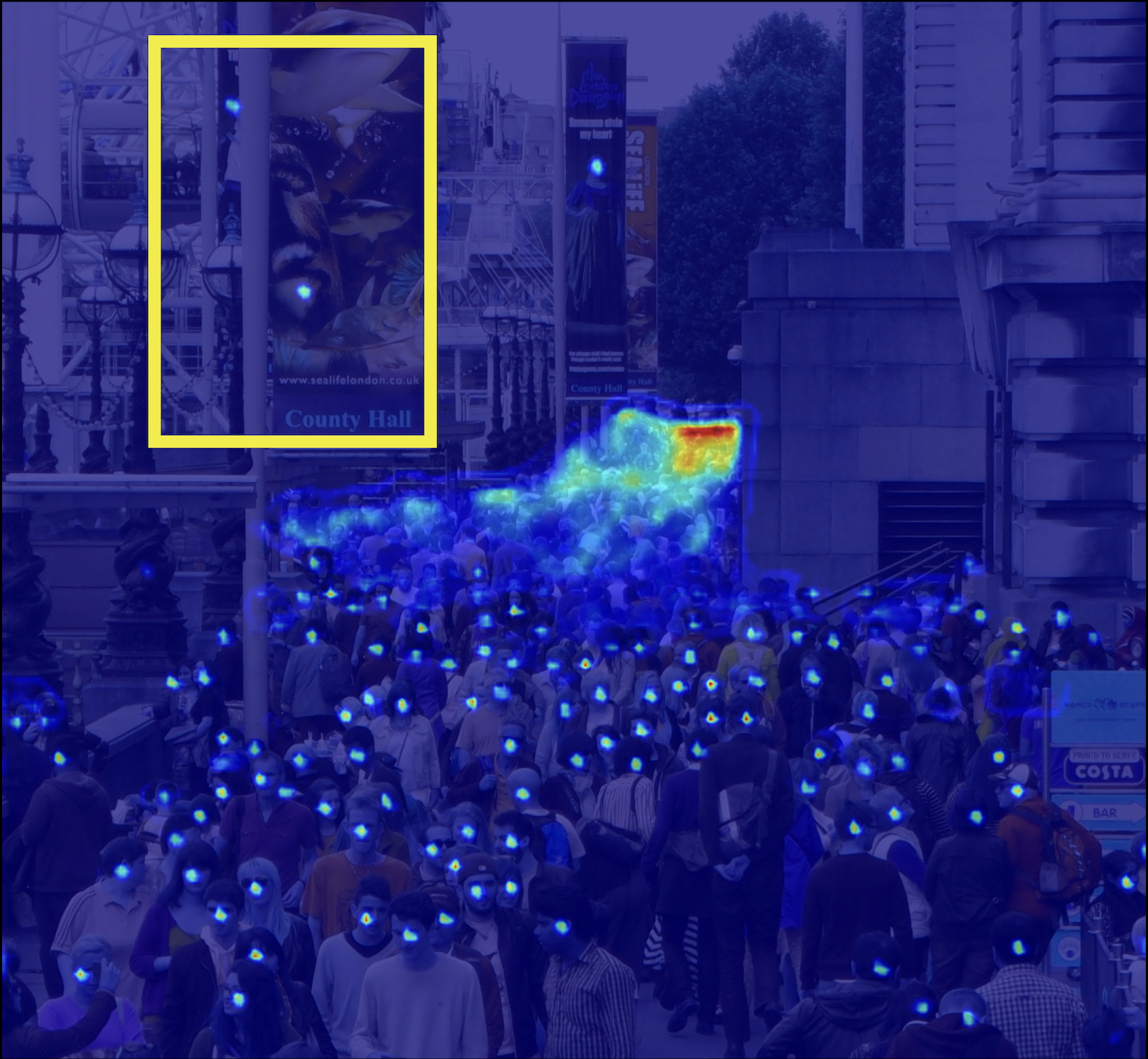} &
			\includegraphics[height=0.18\linewidth]{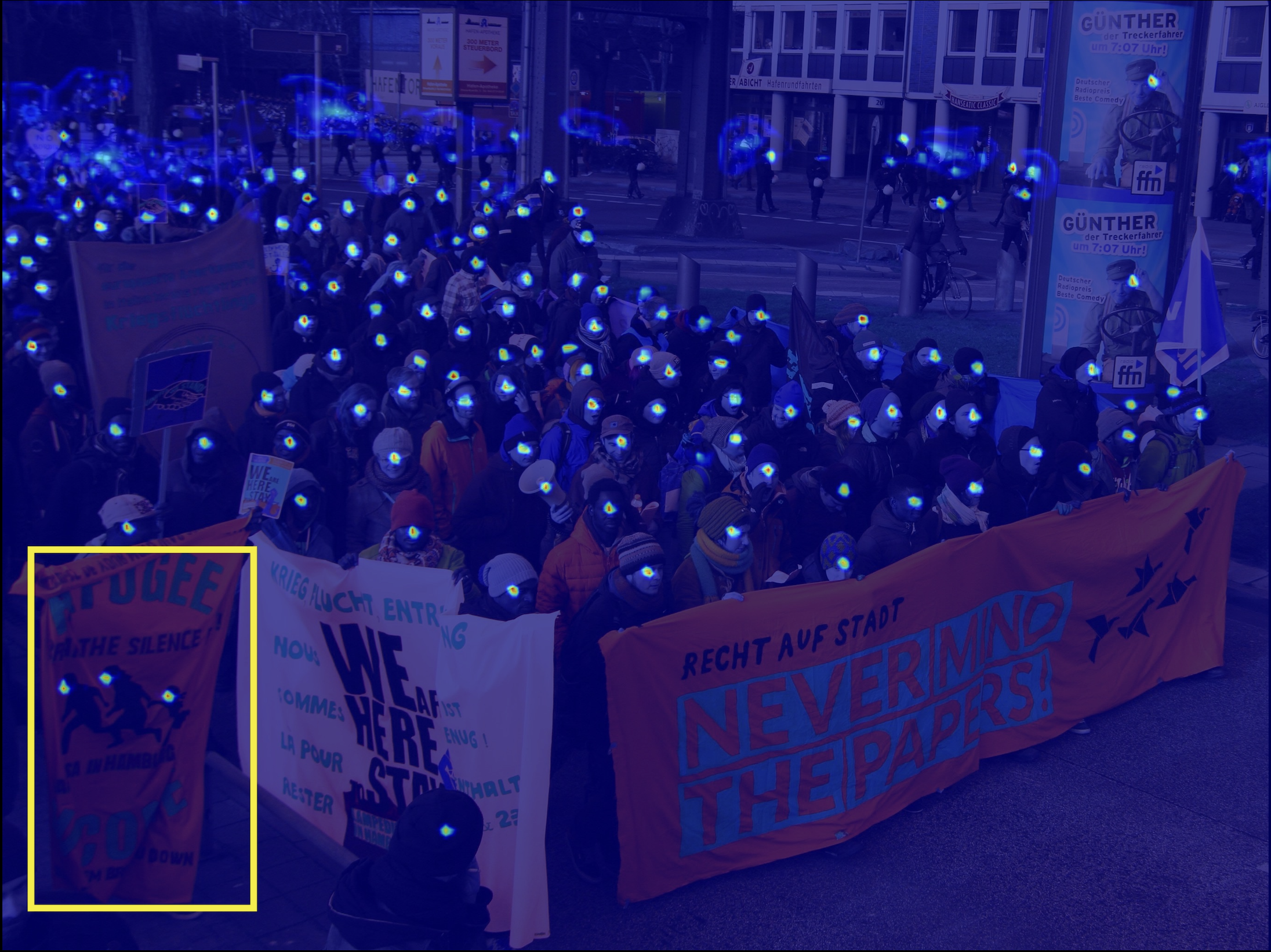} \\

                \includegraphics[height=0.18\linewidth]{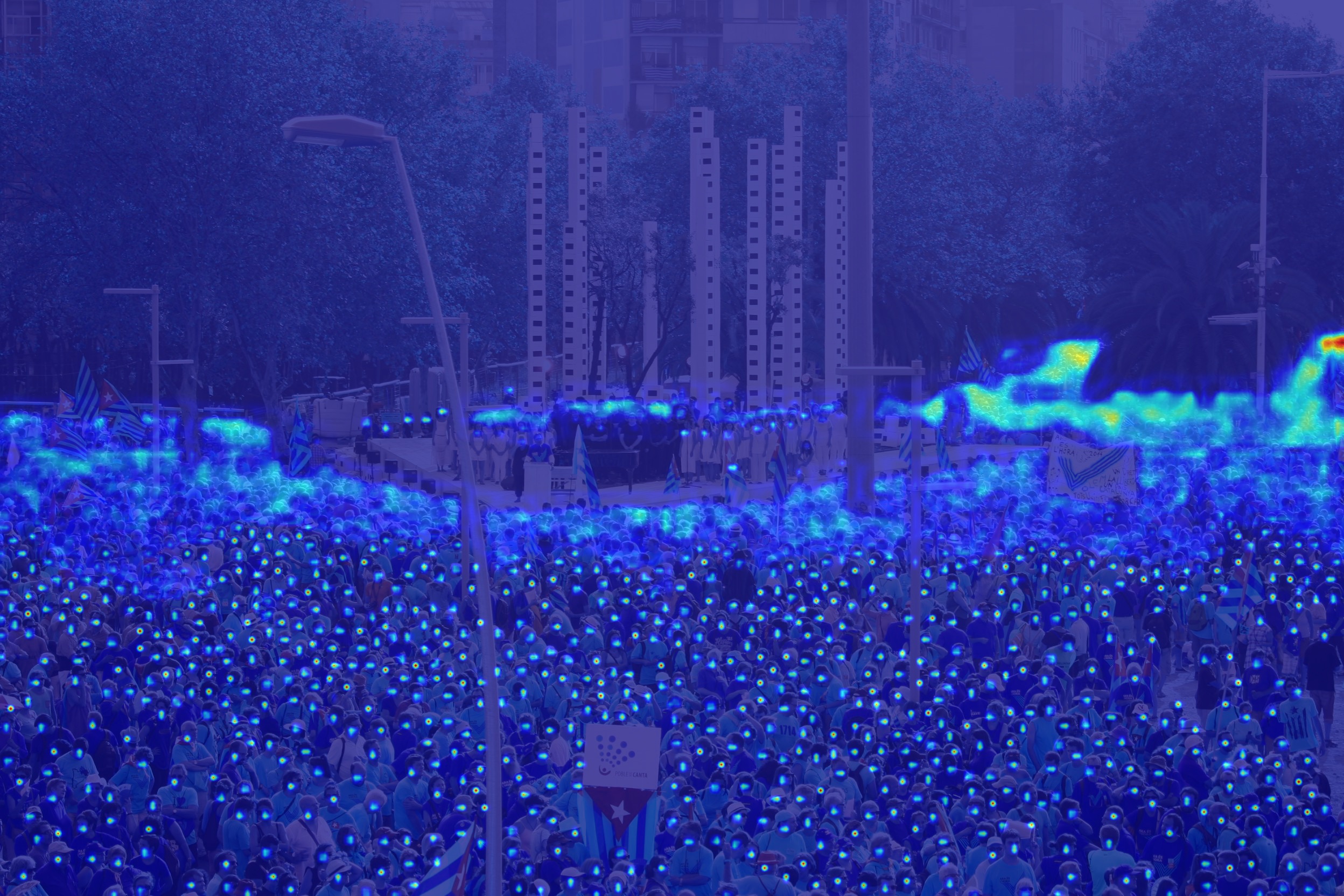}  &
			\includegraphics[height=0.18\linewidth]{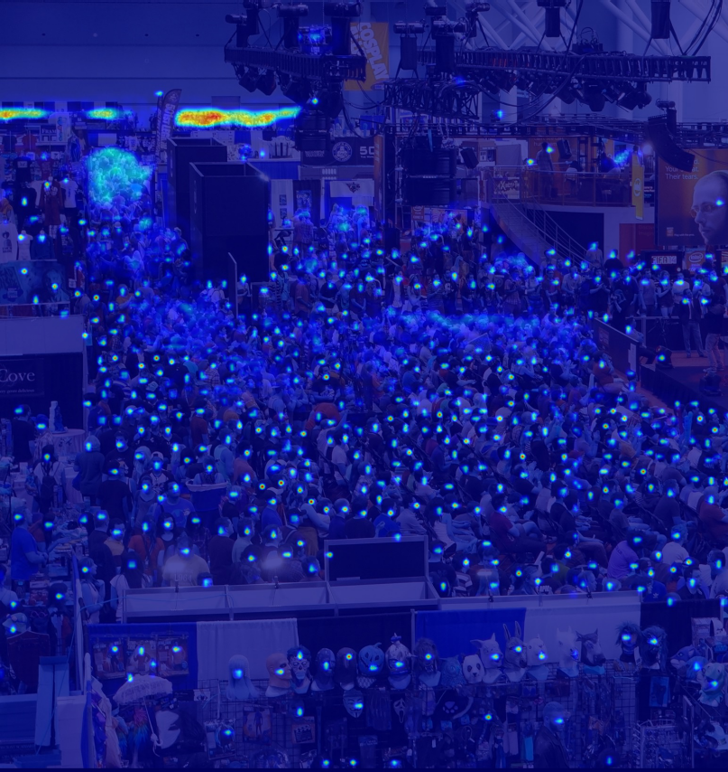} &
			\includegraphics[height=0.18\linewidth]{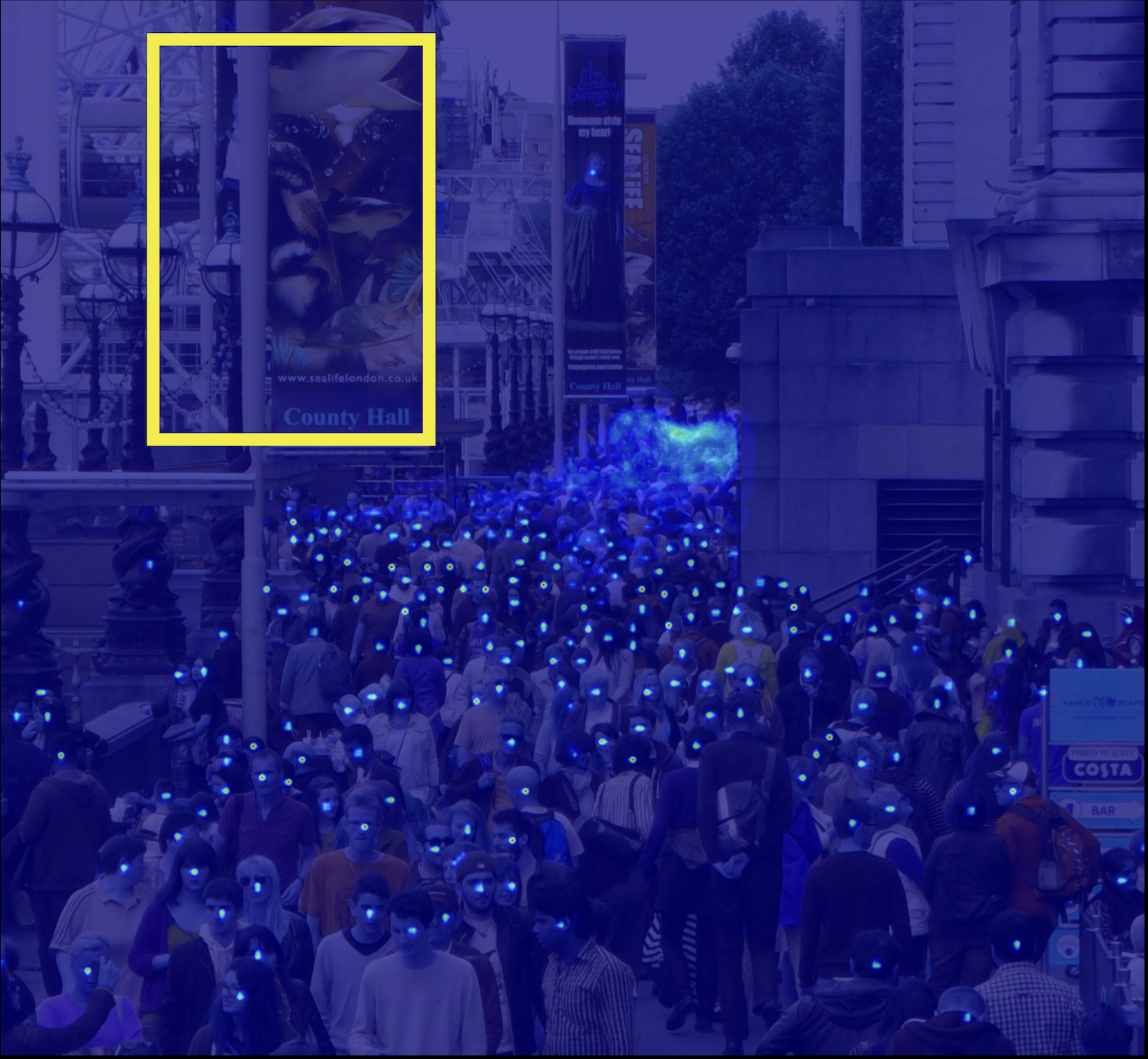} &
			\includegraphics[height=0.18\linewidth]{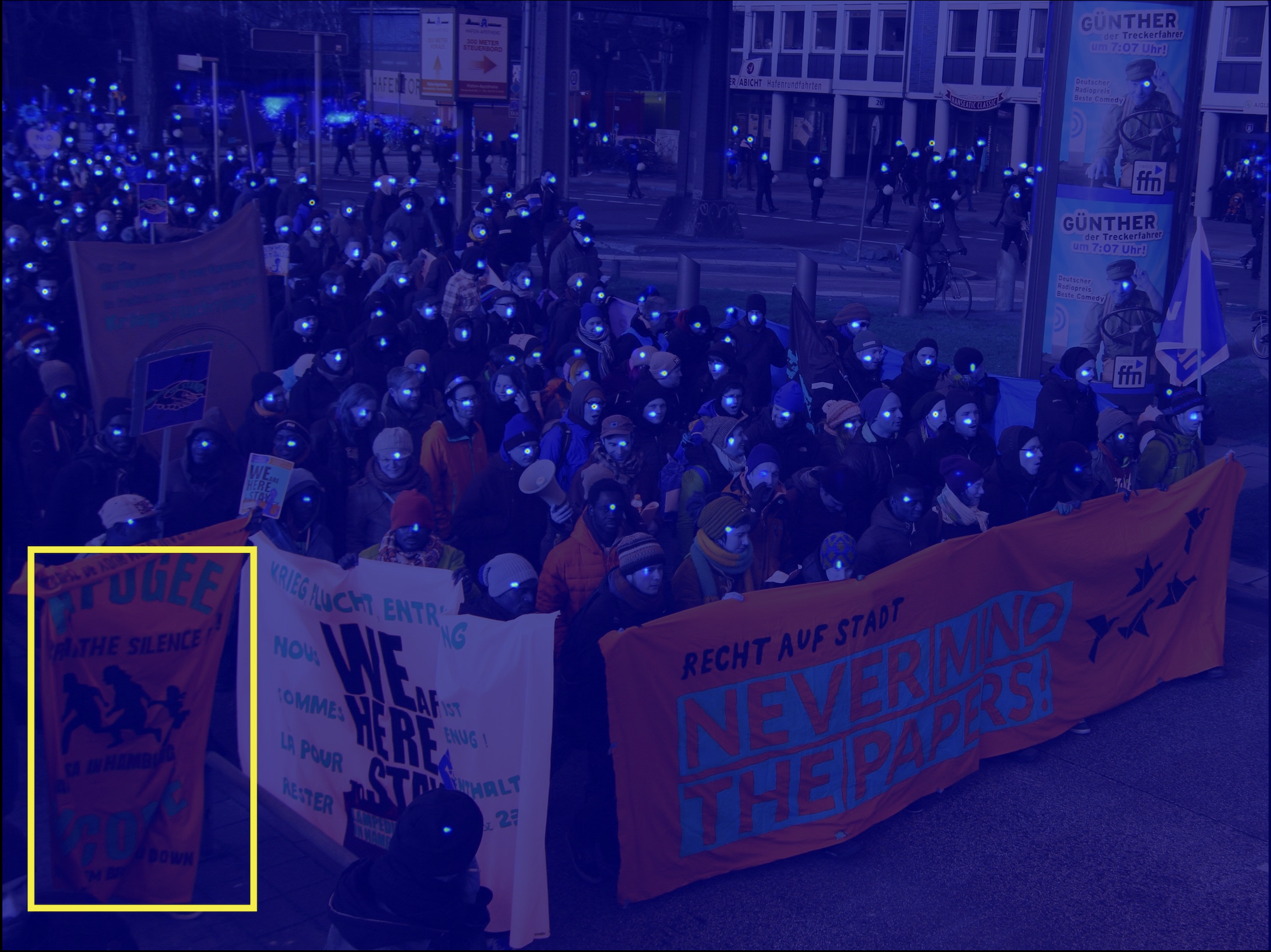} \\

		\end{tabular}
		\caption{Visualizations on UCF-QNRF. The second row contains density maps predicted by the vanilla transformer while the third row contains density maps predicted by our proposed \ourmodel{}. The vanilla transformer predicts blurred densities (the first example) in the crowd with smaller scales, while our method still generates clear density points. For the last two examples, our model avoids false alarms in the background predictions.}
		\label{fig:vis}
	\end{center}
\vspace{-4mm}
\end{figure*}}

\end{document}